\documentclass[conference]{IEEEtran}
\IEEEoverridecommandlockouts
\usepackage{cite}
\usepackage{amsmath,amssymb,amsfonts,subfigure,enumitem,multirow}
\usepackage{algorithmic}
\usepackage{graphicx}
\usepackage[ruled,vlined]{algorithm2e}
\usepackage{textcomp}
\usepackage{dsfont}
\usepackage{xcolor}
\usepackage{caption}
\usepackage{booktabs}
\def\BibTeX{{\rm B\kern-.05em{\sc i\kern-.025em b}\kern-.08em
    T\kern-.1667em\lower.7ex\hbox{E}\kern-.125emX}}
\begin{document}

\newtheorem{thm}{Theorem}[section]
\newtheorem{lem}[thm]{Lemma}
\newtheorem{prop}[thm]{Proposition}
\newtheorem{cor}{Corollary}
\newtheorem{conj}{Conjecture}[section]
\newtheorem{defn}{Definition}[section]
\newtheorem{exmp}{Example}[section]
\newtheorem{rem}{Remark}

\title{A Unified Model for Spatio-Temporal  Prediction Queries with Arbitrary Modifiable Areal Units}


\author{\IEEEauthorblockN{Liyue Chen$^{1,4}$, Jiangyi Fang$^{1,4}$, Tengfei Liu$^{2}$, Shaosheng Cao$^{3*}$, Leye Wang$^{1,4*}$}
\IEEEauthorblockA{$^{1}$ \textit{Key Lab of High Confidence Software Technologies (Peking University)}, \textit{Ministry of Education, China}}
\IEEEauthorblockA{$^{2}$ \textit{China University of Geosciences, Wuhan, China}}
\IEEEauthorblockA{$^{3}$ \textit{DiDi Chuxing, Hangzhou, China}}
\IEEEauthorblockA{$^{4}$ \textit{School of Computer Science, \textit{Peking University}, Beijing}\\
chenliyue2019@gmail.com,shelsoncao@didiglobal.com,leyewang@pku.edu.cn \thanks{*Corresponding author}}
}

\maketitle

\begin{abstract}
Spatio-Temporal (ST) prediction is crucial for making informed decisions in urban location-based applications like ride-sharing. However, existing ST models often require region partition as a prerequisite, resulting in two main pitfalls. Firstly, location-based services necessitate ad-hoc regions for various purposes, requiring multiple ST models with varying scales and zones, which can be costly to support. Secondly, different ST models may produce conflicting outputs, resulting in confusing predictions. In this paper, we propose \textit{One4All-ST}, a framework that can conduct ST prediction for arbitrary modifiable areal units using only one model. To reduce the cost of getting multi-scale predictions, we design an ST network with hierarchical spatial modeling and scale normalization modules to efficiently and equally learn multi-scale representations. To address prediction inconsistencies across scales, we propose a dynamic programming scheme to solve the formulated optimal combination problem, minimizing predicted error through theoretical analysis. Besides, we suggest using an extended quad-tree to index the optimal combinations for quick response to arbitrary modifiable areal units in practical online scenarios. Extensive experiments on two real-world datasets verify the efficiency and effectiveness of \textit{One4All-ST} in ST prediction for arbitrary modifiable areal units. The source codes and data of this work are available at https://github.com/uctb/One4All-ST.
\end{abstract}


\section{Introduction} \label{sec: intro}
With rapid urbanization and advancements in sensing technology, Spatio-Temporal (ST) data with location-based capabilities and timestamps are being widely collected from ubiquitous infrastructure and smart devices. Designing ST models for various tasks (e.g., mobility prediction and abnormal detection) is crucial to empower enterprises and governments to monitor urban dynamics \cite{urban_monitoring_2011,digital_mobility_2023}, manage resources \cite{joint_demand_2021,bike_stg_2022,sthan_2022}, conserve energy \cite{MVSTGN_2023}, and enhance public services \cite{st_aware_2022,roi_demand_2023,ssl_traffic_2023}. This serves as the foundation for making informed decisions.

\begin{figure}[t]
  \centering
  \includegraphics[width=1\linewidth]{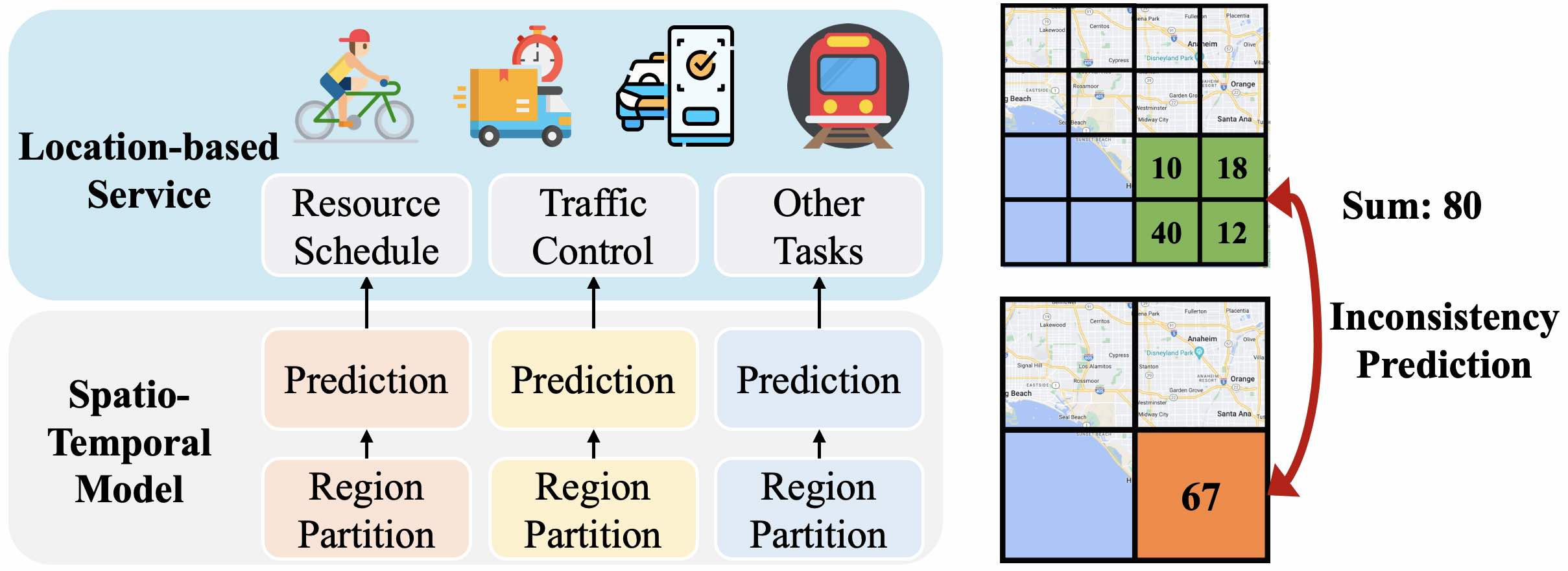}
  \caption{Research motivation. \textbf{Left}: Location-based services require ad-hoc regions for various purposes, necessitating numerous ST models to support the service. \textbf{Right}: The prediction inconsistency raised by different ST models.}
  \vspace{-1.6em}
  \label{fig: motivation}
\end{figure}

Although current ST models perform well with a specific region partition \cite{graph_wavenet_2019, GMAN_2020, MTGNN_2020, STRN_2021, STMeta}, there are still gaps in providing ubiquitous location-based services. Firstly, as shown in Fig.~\ref{fig: motivation}, real-world applications rely on ST prediction with various region specifications as a decision-making basis. For instance, online ride-hailing platforms like Uber require demand prediction tasks and taxi flow control tasks that may involve areas of 1km$^2$ \cite{geng2019spatiotemporal, ST_MetaNet_2022} and 0.25km$^2$ \cite{joint_demand_2021, zheng_deepstd_2020}, respectively. Moreover, prediction tasks may need changes in their analysis areas as urban traffic patterns differ throughout the day \cite{yuan_functions_2012}, leading to corresponding changes in interested community transportation structure \cite{sun_community_2016}. When region specifications change or different scale analyses are needed, relying on many ad-hoc ST models can be costly. Moreover, creating models for each region partition could result in confusing inconsistency which is also famous in geoscience, widely known as the modifiable areal unit problem (MAUP) \cite{wong2004modifiable,openshaw1981modifiable}, since different region specifications may lead to diverged analysis outcomes \cite{de2021multicriteria, GridTuner_2022, RegionGen_2023}. For example, as shown in the right chart of Fig.~\ref{fig: motivation}, coarser models may produce conflicting outputs compared to finer ones, causing confusion about which result to use. 

Hence, a fundamental research question arises: \textbf{can we build a unified ST prediction model for arbitrary Modifiable Areal Units (MAU)?} Such a model can greatly reduce the development and deployment costs and seamlessly adapt to changes in regions of interest over time. However, predicting for arbitrary MAU with just one model is difficult due to variations in scale and zones within the region of interest across different prediction tasks.

One intuitive approach is to create a fine-grained ST model capable of producing coarser results through aggregation. This approach has the advantage of low computational costs since only one model is needed, and pioneering works have made great efforts to develop fine-grained ST models \cite{STRN_2021,fine_inference_incomplete_2023}. However, applying fine-grained ST models to coarser scales may lead to inferior performance. As demonstrated later in our experiments, aggregating the results of the fine-scale \textit{ST-ResNet} \cite{zhang2017deep} model for predictions increases the RMSE (Root Mean Square Error) by 15.2\% compared with directly using predictions of the coarse-scale \textit{ST-ResNet} model.

A more effective yet costly approach is training many ad-hoc models for multi-scale outputs and selecting the appropriate scale for arbitrary region queries\footnote{Without incurring ambiguity, region queries refer to the regions of interest queried by location-based services.}. Recently, a pioneering work \cite{MC_STGCN_2022} simultaneously performs fine- and coarse-grained predictions using a single model through multi-task learning. This approach reduces the number of parameters required for obtaining two-scale predictions, compared to training two separate models. Despite its contribution towards efficient multi-scale ST prediction, challenges remain in providing ubiquitous location-based services with MAU.

\textbf{Challenge 1.} \textit{How to equally learn multi-scale representations in a lightweight manner?} 
Handling modifiable areal units of varying scales requires multi-scale representations and predictions (more than two scales), raising two issues. Firstly, previous models use separate modules to learn multi-scale representation \cite{MC_STGCN_2022}, which can be costly as the number of scales increases. Hence, an efficient structure for multi-scale learning is urgently needed. Secondly, existing multi-task learning methods balance the loss magnitude between multi-scale tasks by manually setting weights \cite{MTGCN_DASFAA_2020, MC_STGCN_2022}. While this method is feasible when the scale structure is relatively simple (e.g., only two scales), it becomes cumbersome and impractical when dealing with a large number of scales.

\begin{figure}[h]
  \centering
  \includegraphics[width=0.9\linewidth]{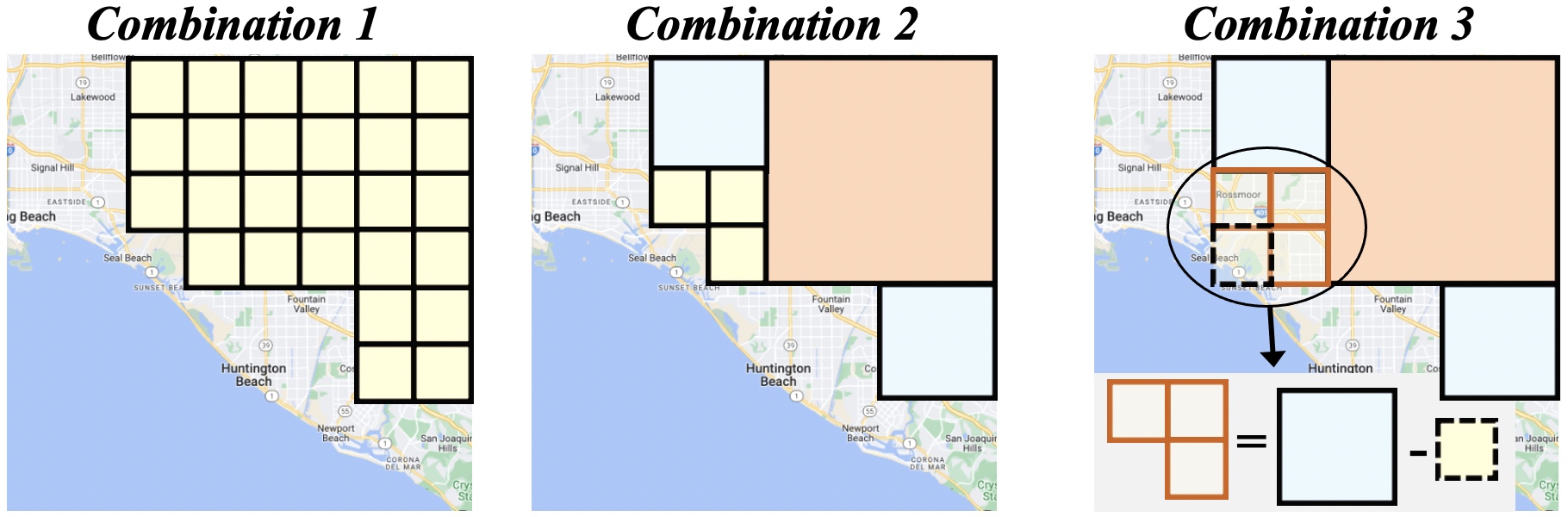}
   \vspace{-0.3em}
  \caption{Three combinations with different predictability for representing the same modifiable areal unit.}
  \vspace{-0.6em}
  \label{fig: exp_comb}
\end{figure}

\textbf{Challenge 2.} \textit{How to effectively represent modifiable areal units using the pre-decided multi-scale regions?} 
The spatial heterogeneity of ST data poses challenges for prediction \cite{STDM_2020}, which can vary across scales and zones \cite{GridTuner_2022, RegionGen_2023}, leading to inconsistencies between predictions for different region specifications. There may exist many feasible combinations to represent modifiable areal units. For example, Fig.~\ref{fig: exp_comb} shows three combinations to represent the same region of interest based on pre-decided grid scales. How do we determine the optimal combination for getting the most accurate predictions?

\vspace{-0.8em}

To address these challenges, we propose a framework called \textit{One4All-ST}. Our main contributions include:
\begin{itemize}[leftmargin=0.4cm]
    \item As far as we know, this is one of the pioneer works studying the ST prediction problem for arbitrary modifiable areal units. It is also a useful attempt to alleviate the expensive cost and prediction inconsistency caused by developing multiple models for location-based service.

    \item \textit{One4All-ST} includes three main components. Firstly, we design a lightweight network with hierarchical spatial modeling and scale normalization modules to efficiently and equally learn multi-scale representations. Secondly, we formulate the optimal combination problem to select appropriate scale outputs for modifiable areal units and solve it with theoretical analysis. Thirdly, we suggest using an extended quad-tree to index the optimal combinations, enabling rapid prediction responses in practical online scenarios.
    
    \item We extensively experiment on two real-world datasets to verify the efficiency and effectiveness of \textit{One4All-ST} in accurately predicting arbitrary modifiable areal units.
\end{itemize}

\section{Preliminaries and Problem Statement} \label{formulation}

\begin{figure}[h]
\centering
\subfigure[Hierarchical grids]{
\includegraphics[width=0.32\linewidth]{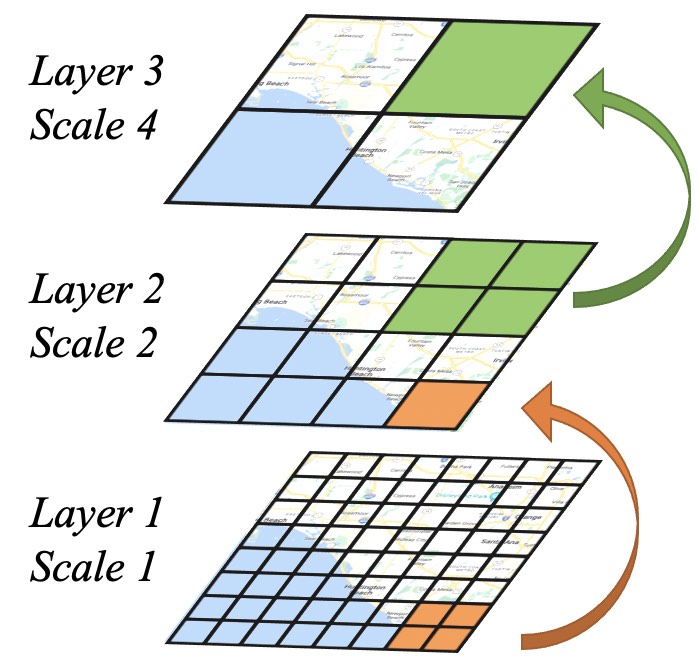}
    \label{hiera_grid}
}\hspace{-3mm}
\subfigure[Rasterized region]{
\includegraphics[width=0.32\linewidth]{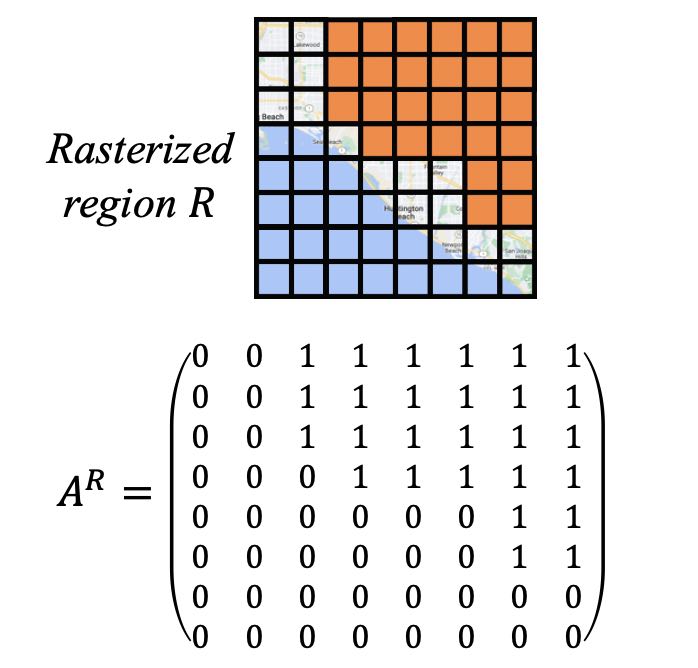}
    \label{road_query}
}\hspace{-3mm}
\subfigure[Example of mapping]{
\includegraphics[width=0.32\linewidth]{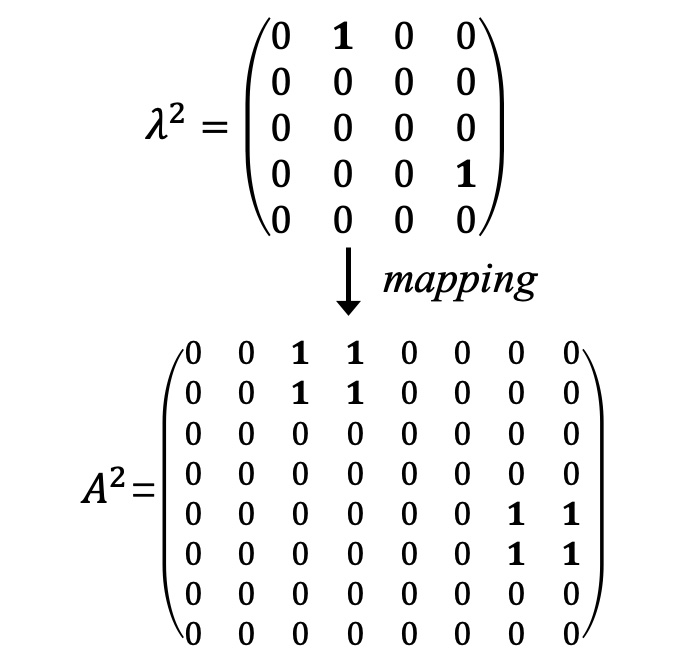}
    \label{exp_mapping}
}
\subfigure[A grid combination of the above region]{
\includegraphics[width=0.92\linewidth]{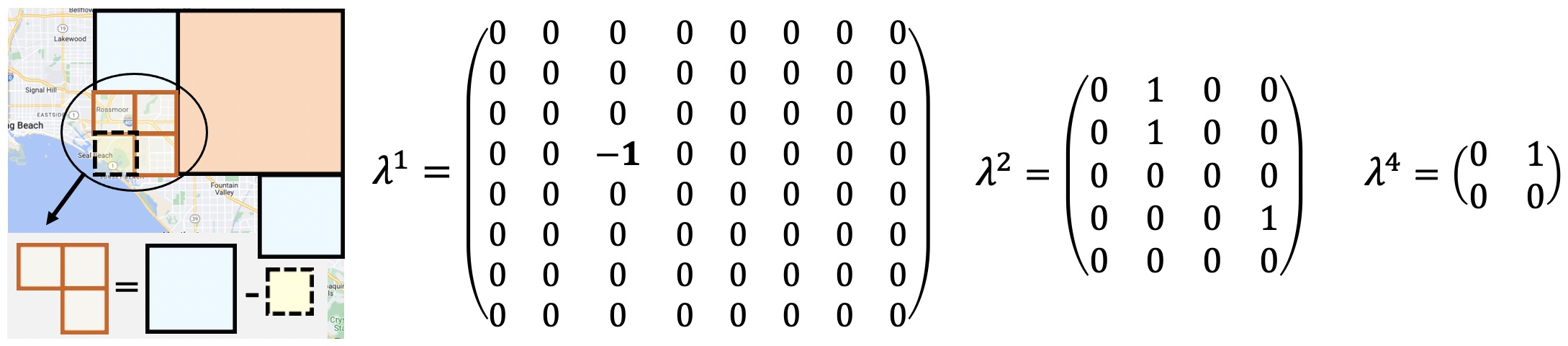}
    \label{comb_demo}
}
\vspace{-.7em}
\caption{(a): Example of hierarchical grids. (b): A rasterized region and its assignment matrix. (c): Example of the mapping function. (d): Example of a grid combination.}
\vspace{-.7em}
\end{figure}

\noindent \textbf{Definition 1 (Hierarchical grids)} We partition an area of interest (e.g., a city) evenly into an atomic raster with totally $N=H\times W$ grids. The atomic raster is in Layer 1 and has the highest resolution (e.g., 150m $\times$ 150m) by setting either the largest $H$ or $W$. Other layers obtain lower-resolution grids by combining adjacent high-resolution grids with a window containing $K \times K$ grids. The stride of sliding windows is also set to $K$, ensuring that each higher-resolution grid belongs to only one lower-resolution grid in a hierarchical structure. Specifically, the window size in Layer $l$ is $K_l$ and Layer $l$ has $H_l \times W_l$ grids, satisfying $H=H_l\cdot \xi_l$ and $W=W_l\cdot\xi_l$ ($\xi_l = \prod_{i=1}^{l-1} K_i$). For convenience, we refer to Layer $l$ that has larger-sized ($\xi_l$ times) grids as Scale $\xi_l$ (abbreviated as $S_{\xi_l}$). Here, $S_{\xi_l}= \mathds{1}_{H_l\times W_l}$ is a matrix where all elements are one. As shown in Fig.~\ref{hiera_grid}, Layer 3 ($S_4$) and Layer 2 ($S_2$) are merged from respective previous layers using a $2\times2$ window.

\begin{figure*}[htbp]
  \centering
  \includegraphics[width=0.9\linewidth]{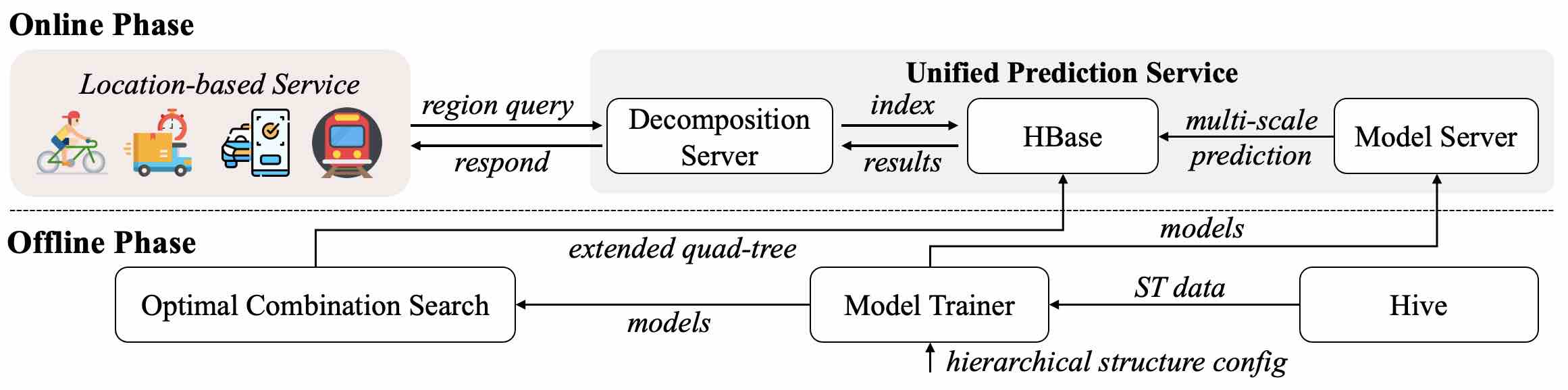}
  \vspace{-0.5em}
  \caption{The workflow of \textit{One4All-ST} system.}
  \vspace{-1.5em}
  \label{fig: systemflow}
\end{figure*}

\noindent \textbf{Definition 2 (Hierarchical structure)} Given $S_1$, which is the atomic raster, and $S_n$, the hierarchical structure $P$ is a set that records all scales contained. For instance, if the size of the window in every layer is $2\times2$, the hierarchical structure from $S_1$ to $S_{16}$ would be $P=\{1, 2, 4, 8, 16\}$.

\noindent \textbf{Definition 3 (Citywide crowd flow \cite{zhang2017deep, STRN_2021})} The crowd flow at time $t$ on Layer $l$ is a 3D tensor $\mathbf{X}^l_t \in \mathbb{R}^{H_l \times W_l \times C}$, where $C$ is the total number of flow measurements (e.g., inflow or outflow). Each entry $(h,w,c)$ indicates the value of the $c$-th measurement in grid $(h,w)$. Specifically, $\mathbf{X}^1_t \in \mathbb{R}^{H \times W \times C}$ denote the citywide crowd flow of atomic grids at time $t$.

\noindent \textbf{Definition 4 (Rasterized region)} The region is a geographic polygon represented by a path that contains a list of geo-coordinates $\langle(lat_1, lng_1),$ $ ...,$ $(lat_n, lng_n)\rangle$ defining its boundaries. Region $R$ can be rewritten as $R = A^R \odot S_1$ by rasterizing and aligning it with atomic grids ($S_1= \mathds{1}_{H\times W}$). The assignment matrix $A^R \in \{0,1\}^{H\times W}$ indicates whether an atomic grid belongs to region $R$ by setting its corresponding value in the matrix to 1 ($A^R_{i,j}=1$). For example, Fig. \ref{road_query} displays a rasterized region and its assignment matrix.

\noindent \textbf{ST Prediction Problem for Modifiable Areal Units} Given a set of arbitrary rasterized regions $\mathcal{R}=\{R_1, R_2,...\}$ derived from various tasks, a series of historical citywide crowd flow time slot 1 to time slot $t-1$ on atomic grids $\{\mathbf{X}^1_1, \mathbf{X}^1_2, ..., \mathbf{X}^1_{t-1}\}$, we want to predict the crowd flow for each rasterized region $R_i$ in the next time slot $t$ to minimize:
\begin{align}
\mathcal{L}(\hat{\mathbf{X}}_{t}^{R_i}, \mathbf{X}_{t}^{R_i})\ \ \ \ \forall R_i \in \mathcal{R}
\end{align}
where $\hat{\mathbf{X}}_{t}^{R_i}$ and $\mathbf{X}_{t}^{R_i}$ are the predicted and ground truth crowd flow of $R_i$ in the next time slot $t$. $\mathcal{L}$ is the loss function (e.g., mean square error). To solve this actual problem, we break it down into the following two sub-problems.

\noindent \textbf{Multi-scale ST Prediction Problem} Given a series of historical citywide crowd flow on atomic grids $\{\mathbf{X}^1_1, \mathbf{X}^1_2, ..., \mathbf{X}^1_{t-1}\}$, the hierarchical structure $P$, we want to predict the crowd flow for each scale $s$ in the next time slot $t$ to minimize:
\begin{align}
\mathcal{L}(\hat{\mathbf{X}}_{t}^{s}, \mathbf{X}_{t}^{s})\ \ \ \ \forall s \in P
\end{align}
where $\hat{\mathbf{X}}^s_{t}$ and $\mathbf{X}^s_{t}$ are the predicted and ground truth citywide crowd flow at Scale $s$ in the next time slot $t$.

\noindent \textbf{Optimal Combination Problem for Modifiable Areal Units} 
Given an arbitrary rasterized region $R$, the hierarchical structure $P$, our target is to find out the optimal combination $\Lambda^{*}(R)$ by minimizing the predicted error:
\begin{align}
    \arg\min_{\Lambda} &= \sum_t \mathcal{L}(\sum_{i\in P}\vert\vert \lambda^s\odot f_{\Theta^{*}}(\mathbf{X}_{t-T:t-1}^s)\vert\vert, \mathbf{X}_{t}^R) \label{comb_opt_problem} \\
    s.t.\ &\Theta^{*}=\arg \min_{\Theta}\sum_{s\in P}\sum_{t}\mathcal{L}(f_{\Theta}(\mathbf{X}_{t-T:t-1}^s),\mathbf{X}_{t}^s) \\
    s.t.\ & \sum_{s\in P} A^{s} = A^R \label{eq: matched_region}
\end{align}
where $\Theta$ is the network parameters. $A^R$ is the assignment matrix of region $R$. 
The combination $\Lambda=\{\lambda^s\vert s\in P\}$ includes assignment matrices at all scales within the hierarchical structure $P$. Here, $\lambda^s$ represents the assignment matrix at Scale $s$. 
$\lambda^s$ can be converted into atomic grids represented as $A^s \in \{-1,0,1\}^{H\times W}$ using the mapping function $A^s_{i,j}=\lambda^s_{\lfloor \frac{i}{s}\rfloor,\lfloor \frac{j}{s}\rfloor}$. Fig.~\ref{exp_mapping} provides an illustrative example of this mapping process. 
In $\lambda^s$, a value of `1' and `-1' indicates that we should take this grid into consideration by union or subtraction. Fig.~\ref{comb_demo} illustrates an example combination involving both union and subtraction operations. The sum of combinations across all scales must equal the rasterized region (Eq. \ref{eq: matched_region}).

\section{System Workflow} \label{system_workflow}
In this section, we elaborate on the workflow of the proposed \textit{One4All-ST} system. As illustrated in Fig. \ref{fig: systemflow}, our system comprises two stages: offline phase and online phase.

In the offline phase, our system initiates by training a multi-scale spatio-temporal network using ST data stored in Hive~\cite{Hive}. 
Our analysis shows that for arbitrary MAU, the optimal combination is achieved through aggregating the optimal combinations of decomposed hierarchical grids (Theorem \ref{thm_1}).
Therefore, with the trained models to assess the quality of combinations, our system searches for the optimal combination for all grids within the pre-decided hierarchical structure.
Then, our system constructs a quad-tree to index the optimal combinations of all grids and transmits the index to HBase~\cite{HBase}, ensuring swift responses for online predictions.

In the online phase, the deployed ST model continuously synchronizes multi-scale predictions with HBase at preset intervals. The region decomposition server receives region queries from the location-based services and will decompose them into grids with various scales. Then, the server retrieves the optimal combination for every grid based on the index and obtains the final prediction by aggregating all grids. On our experiment platform (Sec.~\ref{implementation_details}), our system's response time for each region query is within 20 milliseconds (Sec.~\ref{sec: response_time}), which is sufficient for online services.


\section{Methodology}
\subsection{Framework Overview}
\begin{figure}[h]
  \centering
  \includegraphics[width=1\linewidth]{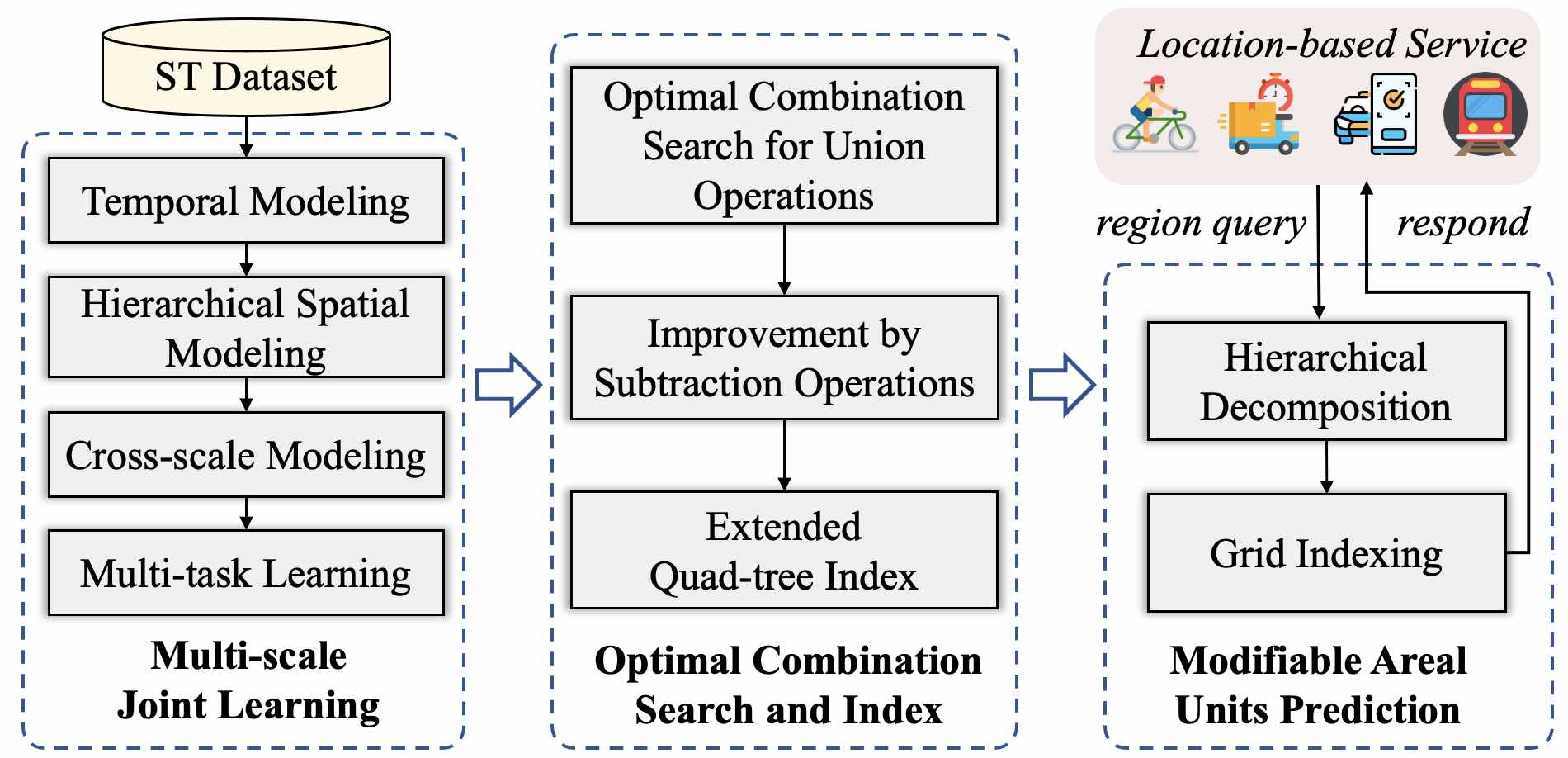}
  \caption{Overall framework of \textit{One4All-ST}.}
  \label{fig: framework}
  \vspace{-1.5em}
\end{figure}

\begin{figure*}[htbp]
  \centering
  \includegraphics[width=0.8\linewidth]{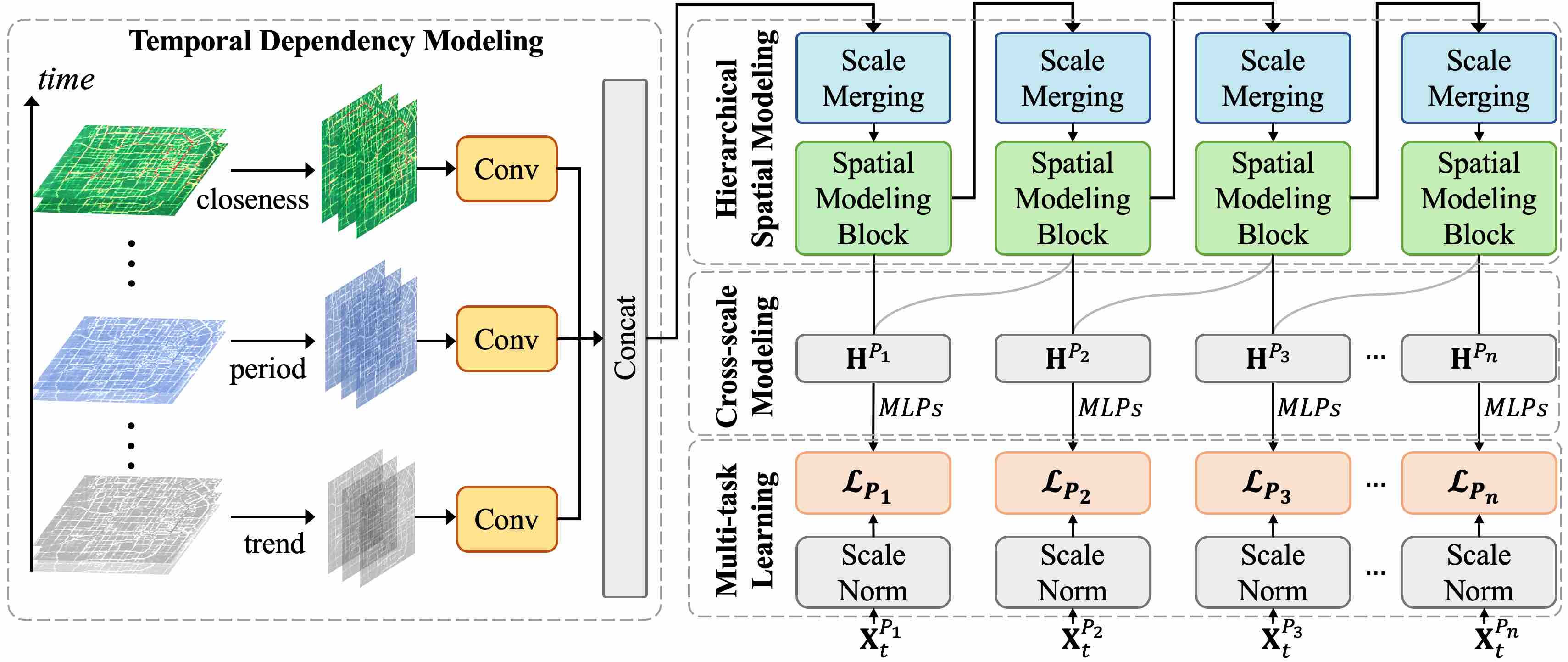}
  \caption{The proposed hierarchical multi-scale spatio-temporal network (i.e., the multi-scale joint learning component).}
  \vspace{-1em}
  \label{fig: network}
\end{figure*}

The proposed framework called \textit{One4All-ST}, has three components: multi-scale joint learning, optimal combination search and index, and modifiable areal units prediction (Fig.~\ref{fig: framework}). 

In the multi-scale joint learning component, we propose a hierarchical ST network for multi-scale predictions, integrating a temporal modeling module to capture temporal dependencies and a stacked hierarchical structure for efficient spatial representation learning. Additionally, the cross-scale modeling module enhances ST representations by leveraging information from other scales, all of which are then fed into the multi-task learning module for multi-scale learning.

In the optimal combination search and index component, we first show that the optimal combination of modifiable areal units can be achieved by aggregating optimal combinations of decomposed hierarchical grids through union operations. We employ a dynamic-programming search approach to find the optimal combinations for each hierarchical grid. Then, we also explore subtraction operations to find other feasible and improved combinations. After completing these searches, we construct a quad-tree to index the optimal combinations, accelerating online prediction for modifiable areal units.

The modifiable areal units prediction component is designed for online use. It first decomposes region queries from location-based services into hierarchical grids. The final prediction is derived by aggregating these grids utilizing optimal combinations indexed from the pre-established quad-tree.

\subsection{Multi-scale Joint Learning} \label{learning}
The multi-scale joint learning component aims to efficiently make multi-scale predictions. Our proposed hierarchical multi-scale ST network (Fig.~\ref{fig: network}) comprises four modules: \textit{Temporal Modeling}, \textit{Hierarchical Spatial Modeling}, \textit{Cross-scale Modeling}, and \textit{Multi-task Learning}.

Compared with previous multi-scale ST networks, our improvement lies in the following two aspects. Firstly, our hierarchical spatial modeling module learns the spatial representations hierarchically by stacking the spatial modeling block layer by layer instead of using a totally different spatial modeling block for each scale as in previous research \cite{HiSTGNN_2022, MC_STGCN_2022}. This approach is more efficient for deep hierarchical structures (as later shown in our experiment, our model achieves better accuracy with six scales and half the parameters compared to previous work that only used two scales \cite{MC_STGCN_2022}) since it extracts the spatial representation of coarser grids from finer grids in previous layers rather than extracting them from scratch. Secondly, our multi-task learning module incorporates a scale normalization mechanism to balance the learning tasks of different scales and ensure balanced consideration for every scale. This approach is more reasonable than manually assigning task weights done in previous studies \cite{MTGCN_DASFAA_2020, MC_STGCN_2022}.

\subsubsection{Temporal Modeling}
To capture different temporal dependencies, previous research selects several slots (closeness, period, and trend) along the time \cite{zhang2017deep} and this paradigm is widely proven efficient in related research \cite{STRN_2021, STMeta}. Following this, we select recent, near, and distant atomic rasters to predict the citywide crowd flow at $t$:
\begin{equation}
\begin{aligned}
\mathbf{XC}_{t} &= [\mathbf{X}_{t-l_c}^{1},\mathbf{X}_{t-(l_c-1)}^{1},...\mathbf{X}_{t-1}^{1}] \\
\mathbf{XP}_{t} &= [\mathbf{X}_{t-l_d*d}^{1},\mathbf{X}_{t-(l_d-1)*d}^{1},...\mathbf{X}_{t-d}^{1}] \\
\mathbf{XT}_{t} &= [\mathbf{X}_{t-l_w*d}^{1},\mathbf{X}_{t-(l_w-1)*w}^{1},...\mathbf{X}_{t-w}^{1}]
\end{aligned}
\end{equation}
where $d,w$ are the daily and weekly intervals respectively (e.g., in a one-hour prediction task, $d$ and $w$ are 24 and 144). We use three non-shared convolutional layers to convert them to temporal representations, each with $D$ channels, i.e., they are all in $\mathbb{R}^{H\times W \times D}$. Next, we concatenate these three temporal features and get the fused representations at Scale 1.
\begin{equation}
\mathbf{h}_{t}^1 = \texttt{Concat}(\texttt{Conv}(\mathbf{XC}_{t});\texttt{Conv}(\mathbf{XP}_{t});\texttt{Conv}(\mathbf{XT}_{t}))
\end{equation}

\subsubsection{Hierarchical Spatial Modeling}
The spatial modeling design for all scales includes a scale merging layer and a spatial modeling block.
The scale merging layers aggregate adjacent grids, reducing the width and height of feature maps. This process maps spatial representations from finer-grained grids to coarse-grained ones. The subsequent spatial modeling block captures spatial dependencies and learns the representation for a specific scale. Suppose that the hierarchical structure $P=\{P_1,P_2,...P_n\}$ has $n$ scales, the spatio-temporal representations at time $t$ are computed by:
\begin{equation}
    \mathbf{h}^{P_i}_t = \texttt{SM}(\texttt{Merge}(\textbf{h}^{P_{i-1}}_t)),\ \ 2 \leq i \leq n
\end{equation}
where $\texttt{SM}(\cdot), \texttt{Merge}(\cdot)$ are the spatial modeling block and scale merging layer respectively. $P_i$ is a natural number representing the scale, where, for instance, $P_1$ always equals 1, signifying the atomic raster (Scale 1) with $H\times W$ grids. From Scale 1 (i.e., $P_1$) to Scale $P_n$, by adding more scale merging layers and spatial modeling blocks, the learned multi-scale spatial representations are denoted as $\{\mathbf{h}^{P_1}_t,\mathbf{h}^{P_2}_t,...,\mathbf{h}^{P_n}_t\}$.

\textbf{Scale Merging Layer}. In layer $l-1$, suppose we have a $H^{\prime}\times W^{\prime}\times F$ feature map at time $t$, with a merging window size of $K\times K$. The scale merging layer takes this feature map as input, concatenates features within each group of $K \times K$ neighboring grids, and applies a linear layer to reduce the $K^2 \times F$-dimensional features back to $F$ channels. The scale merging layer downsamples the resolution by a factor of $K\times K$, resulting in an output feature map sized $\frac{H^{\prime}}{K} \times \frac{W^{\prime}}{K}\times F$. The scale merging layer can be easily implemented using a standard 2D convolutional layer with kernel size equal to $K$ and stride equal to $K$ (i.e., $\texttt{Merge}(\cdot)=\texttt{Conv}(\cdot)$).

\begin{figure}[h]
  \centering
  \includegraphics[width=0.95\linewidth]{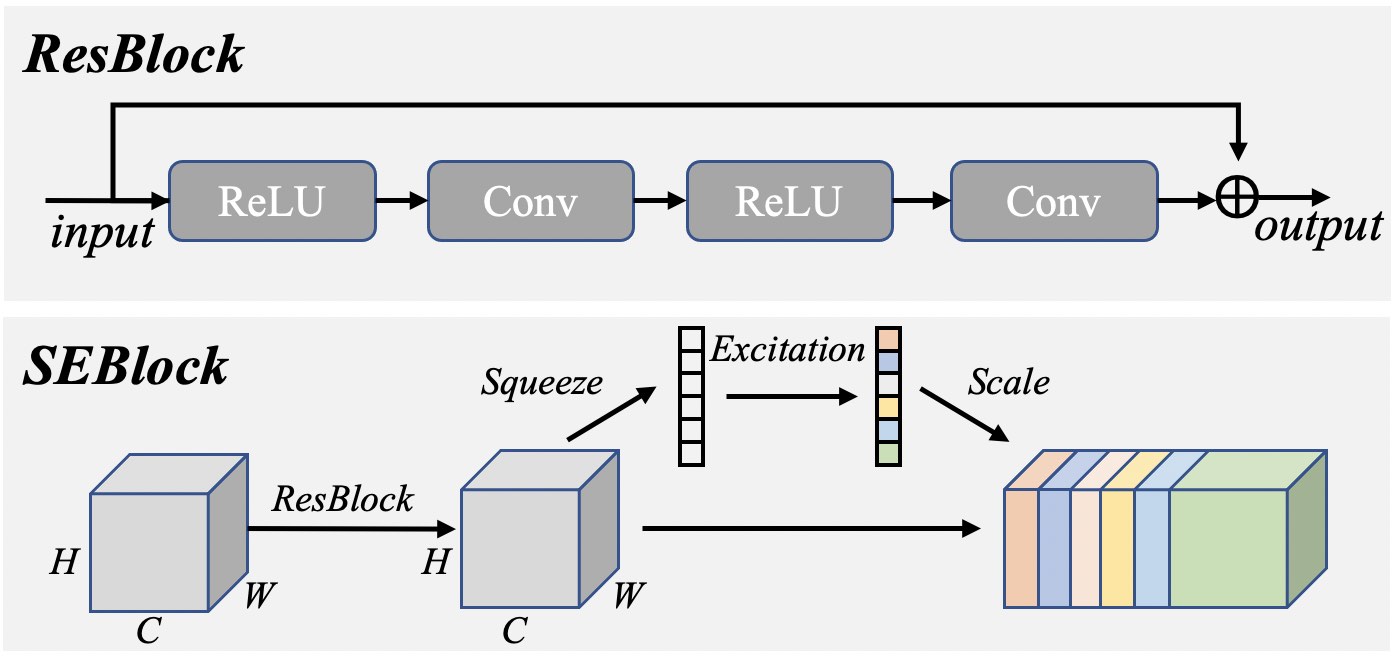}
  \vspace{-0.3em}
  \caption{ResBlock and SEBlock.}
  \vspace{-1em}
  \label{fig: seblock}
\end{figure}

\textbf{Spatial Modeling Block}.
A powerful spatial modeling block is essential for the model to learn discriminative spatial representations. Popular spatial modeling techniques such as ConvBlock \cite{deepst_2016} (i.e., standard convolution block), ResBlock \cite{zhang2017deep}, and SEBlock \cite{STRN_2021} have been widely used in previous ST prediction models. Swin-Transformer \cite{swinTrans_2021} has recently achieved great success in ST modeling \cite{bi2023accurate} and can also be applied for spatial modeling. 
In this paper, we follow the previous work \cite{SENet2018, STRN_2021} by using squeeze-and-excitation (SE) blocks (Fig.~\ref{fig: seblock} illustrates the architecture of SEBlock).

\vspace{-0.8em}

SEBlock leverages spatial and channel-wise information within local receptive fields at each layer without incurring too many training parameters like attention-based methods do. We believe the choice of spatial modeling block is a trade-off between efficiency and performance. In practice, developers can actually select their own preferred spatial modeling blocks, and we also test some other alternatives, like ConvBlock and ResBlock, in our experiments.

\subsubsection{Cross-scale Modeling}
\begin{figure}[h]
  \centering
  \includegraphics[width=0.95\linewidth]{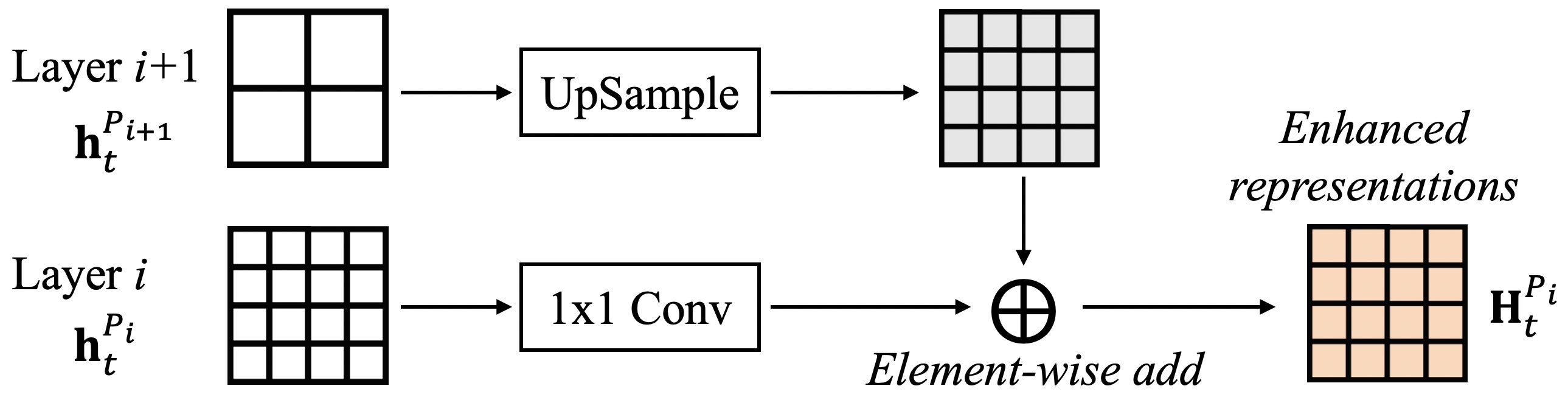}
  \caption{Illustration of the cross-scale modeling.}
  \vspace{-1em}
  \label{fig: cross_scale_demo}
\end{figure}

Previous studies have revealed that coarse-scale representations can benefit fined-grained scale predictions \cite{STRN_2021, MC_STGCN_2022}, making cross-scale modeling necessary. In these studies, coarser scales are irregular and typically use inverse mapping from small regions to large regions to align the feature maps. Given our grid's regular hierarchical structure, with upper grids consistently being integer multiples of lower grids, we can efficiently obtain coarse-scale representations from finer scales using a bottom-up pathway, which is similar to the feature pyramid network \cite{FPN_2017}. Borrowing the idea from feature pyramids in computer vision \cite{FPN_2017}, we adopt a top-down lateral connection to enhance the ST representations, which can be expressed as:
\begin{equation}
    \mathbf{H}^{P_i}_t = \textbf{h}^{P_i}_t + \texttt{UpSample}(\textbf{h}^{P_{i+1}}_t),\ \ 1 \leq i \leq n-1
\end{equation}
Fig.~\ref{fig: cross_scale_demo} gives an illustrative example of cross-scale representation enhancement.
The function $\texttt{UpSample}(\cdot)$ aligns feature maps from coarse scales with finer feature maps by increasing spatial resolution through nearest neighbor upsampling. Subsequently, the lateral connection merges feature maps with the same resolution from both the bottom-up and top-down pathways via element-wise addition. This iterative process starts from the coarsest scale and concludes at the finest scale.

\vspace{-0.5em}


\subsubsection{Multi-task Learning}
After obtaining spatio-temporal representations for each layer, we feed them into multiple fully connected layers to obtain final multi-scale predictions. It is worth noting that these fully connected layers are scale-specific and will not share parameters across different scales.
\begin{equation}
    \hat{\mathbf{X}^{s}_t} = \texttt{MLPs}(\mathbf{H}^{s}_t),\ \ s\in P
\end{equation}
Based on predictions ($\hat{\mathbf{X}^{s}_t}$) and ground truths ($\mathbf{X}_{t}^s$) at every scale, we can train the multi-scale network by combining the losses of different scales. However, a notable challenge arises from significant differences in loss scale attributed to diverse prediction targets across the scales. For instance, crowd flow on the coarsest scale may be over 1,000 times greater than that on the finest scale. These disparities may bias the optimization of the multi-task loss toward coarse scales, resulting in suboptimal performance on fine-grained scales.

To address this, prior studies try to balance learning tasks by manually assigning different weights to each task (e.g., setting a smaller weight for the coarse scale \cite{MC_STGCN_2022}). However, this manual weight assignment is hardly set optimally due to our relatively deep hierarchical structure that typically consists of 5 or 6 scales. Instead of balancing tasks at the loss level, we propose an adaptive normalization mechanism that conducts normalization transformation at the input level for each scale. Specifically, the normalized input for each scale is:
\begin{equation}
    \tilde{\mathbf{X}}^s = \frac{\textbf{X}^s-\mathbb{E}[\mathbf{X}^s]}{\sqrt{\text{Var}[\textbf{X}^s]}},\ \ s \in P
\end{equation}
Then, both the inputs and their losses across all scales are rescaled to a consistent magnitude. This approach ensures equal consideration for each learning task across different scales. As a result, the multi-task loss can be computed straightforwardly as the sum of each scale's loss without the need to set hardly-tuned sum weights \cite{MC_STGCN_2022}:
\begin{equation}
    \min_{\Theta} \sum_{s\in P}\sum_{t}\mathcal{L}(\hat{\mathbf{X}^s_t},\tilde{\mathbf{X}}_{t}^s)
\end{equation}
where $\mathcal{L},\Theta$ are the loss function (e.g. mean square error) and network parameters. $P$ is the hierarchical structure consisting of the scales used (e.g., $P=\{1, 2, 4, 8, 16\}$).

\subsection{Optimal Combination Search and Index} \label{optimal_comb}

Given girds with different scales, one intuitive way to get arbitrary modifiable areal units is union operations, which is a spatial operation that combines the geometries to create a new geometry that represents the spatial extent in geographic information system \cite{spatial_trans_vis} and also widely integrated into GIS tools likes ArcGIS~\cite{union_analysis}. The union operation follows the concept of set-theoretic amalgamation and will output a new geometry that covers the combined area of the input geometries. We refer to the system of modifiable areal units that can be combined by a union of grids of different scales as the union system.

Eq. \ref{comb_opt_problem} describes the objective of the combination optimization problem that we need to find out the optimal hierarchical grid combination for an arbitrary area with the lowest prediction errors. This problem is intractable since there are many feasible grid combinations for $R$. Besides, getting the optimal combination by brute force is time-consuming and infeasible for the online system providing real-time spatio-temporal prediction. 

\begin{figure}[h]
  \centering
  \includegraphics[width=0.99\linewidth]{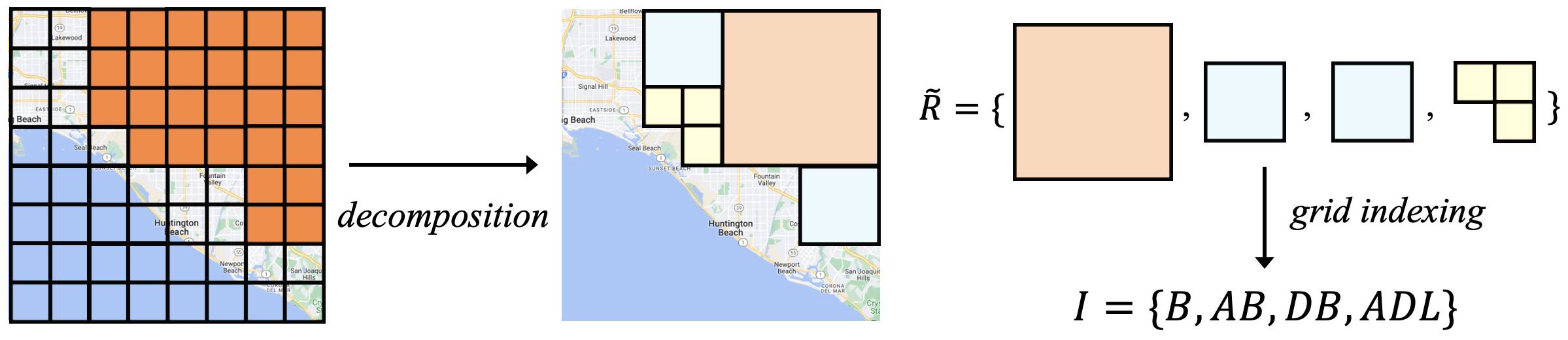}
  \vspace{-1.5em}
  \caption{Illustration of the hierarchical region decomposition.}
  \vspace{-0.5em}
  \label{fig: decomposition}
\end{figure}

To address this issue, we first prove that, given a union system (i.e., only union operations can be used), the optimal grid combination for an arbitrary areal unit can be achieved by aggregating optimal combinations of decomposed fine-grained hierarchical grids. This assertion holds in the absence of further feasible combinations between hierarchical grids. In pursuit of this, we introduce a method (Algorithm \ref{alg: decomposition}) dedicated to decomposing modifiable areal units into hierarchical grids. Its principal design is to decompose in a coarse-to-fine manner, preventing the decomposed grids from being merged into any coarser hierarchical grids. This ensures that there are no additional feasible combinations between the decomposed grids. Fig.~\ref{fig: decomposition} shows an example of hierarchical decomposition.

With this property, we only need to search for the optimal combination for every preset-scale hierarchical grid, whose search space is much smaller and can be done in an offline manner. The following theorem ensures its effectiveness.

\begin{thm} \label{thm_1}
Given a union system, for any rasterized region $R$, which can be decomposed into a set of fine-grained grids $\tilde{R}=\{r_1,r_2,...,r_m\}$ (by Algorithm \ref{alg: decomposition}). We have $\Lambda^*(R) = \Lambda^*(r_1)+\Lambda^*(r_2)+...+\Lambda^*(r_m)$.
\end{thm}

\textit{Proof.} Decompose $R$ into hierarchical grids $\tilde{R}$ that do not intersect and cannot be merged into coarser ones, indicating that there are no other feasible combinations. $\blacksquare$

\vspace{-0.2em}
\SetKwFunction{FMain}{Match}
\SetKwProg{Fn}{Function}{:}{}

\begin{algorithm}[h]
\small
\caption{Hierarchical Decomposition}\label{alg: decomposition}
\LinesNumbered
\KwIn{An arbitrary rasterized region $R$, a set of hierarchical grids $\{S_{P_1},...,S_{P_n}\}$ in $P$ ($\vert P\vert =n$)}
\KwOut{The decomposed hierarchical grids set $\tilde{R}$.}
Initialize $\tilde{R} \gets \emptyset$ \;

\For{$S \gets S_{P_n},S_{P_{n-1}},...,S_{P_1}$}{
    $\mathcal{B} \gets$ \FMain{$R$, $S$} \;
    \For{each $b \in \mathcal{B}$}{
    append connected grids $b$ to $\tilde{R}$ \;
    $R \gets R - b$ \;
    }
    }
\Return $\tilde{R}$
\\\hrulefill \\
\Fn{\FMain{$R, S$}}{
Initialize $V \gets \emptyset$ \;
\For{each grid $s \in S$}{
    \If{$s \subseteq R$}{
    append grid $s$ to $V$ \;
    }
}
Create a graph $G$ using the grid in $V$ as nodes \;
\For{each pair of nodes $(u,v) \in G$}{
    \If{$u,v$ share the same upper grid and are adjacent}{
        connect the edge between $u$ and $v$;
        }
$\mathcal{B} \gets \mathtt{Connected Components}(G)$ \;
}
\Return $\mathcal{B}$
}
\end{algorithm}
\vspace{-0.3em}

\subsubsection{Optimal Combination Search with Dynamic Programming for Union Operations} \label{optim_search_union}
Theorem~\ref{thm_1} reveals that to get the optimal combinations for any given region $R$, we just need to search for the optimal combination in each hierarchical grid.

For coarse hierarchical grids, potential candidate combinations include either utilizing the grids directly or aggregating finer grids with hierarchical relationships through union operations. In general, assuming the merging window size $K$ is a constant, for the hierarchical grids in layer $l$ ($l\ge2$), there are $\sum_{i=1}^{l-1} \frac{N}{K^{2i}}$ ($N=H\times W$) potential combinations from the first layer (i.e., atomic raster) to layer $l-1$. Therefore, for a $n$-layer hierarchical structure, the number of searches is:
\begin{equation}
\sum_{l=2}^n\sum_{j=1}^{l-1} \frac{N}{K^{2j}}= N(\frac{n-1}{K^2-1}-\frac{1-K^{2-2n}}{(K^2-1)^2})
\end{equation}
This results in a time complexity of $O(HWn)$ for the search.

\begin{lem} \label{search_order}
In the hierarchical structure, we can find the optimal combinations of grids in layer $l$ by searching only the grids in layer $l-1$, once we know their optimal combinations.
\end{lem}

\textit{Proof}. Let us consider a case in Fig.~\ref{fig: quadtree}, which applies to other layers or grids as well. Grid $A$ (described by the quad-tree index in Sec.~\ref{quadtree}) is located in Layer 3. The optimal combination of Grid $A$ is $\Lambda^*(A)$, which has the following possible situations from Layer 1 to Layer 3.
\begin{align*}
\Lambda^*(&A)\in\{A,\\
&AA+AB+AC+AD,\\
&AAA+AAB+AAC+AAD+AB+AC+AD,...\}
\end{align*} 
We have found that $\Lambda^*(AA)$ is the better choice between $AA$ and $AAA+AAB+AAC+AAD$. Additionally, $\Lambda^*(AB)$, $\Lambda^*(AC)$, and $\Lambda^*(AD)$ also have searched for the optimal combination with the grids in the lower layer. Therefore, we can determine that:
$\Lambda^*(A)\in\{A,\Lambda^*(AA)+\Lambda^*(AB)+\Lambda^*(AC)+\Lambda^*(AD)\}$.
This means that we can find the optimal combinations of a grid in layer $l$ by only searching for optimal combinations in layer $l-1$. $\blacksquare$

Lemma \ref{search_order} shows that the optimal combination of coarse grid (i.e., $\Lambda^*(A)$) can be constructed from the optimal combination of its sub-problem (e.g., $\Lambda^*(AA)$). Hence, we employ a bottom-up dynamic-programming schedule, facilitating the search by traversing from the finest layer to the coarsest layer in a single pass. As a result, in the union system, the time complexity of searching optimal combinations for hierarchical grids is reduced from $O(HWn)$ to $O(HW)$.

\subsubsection{Improvement by Subtraction Operations} \label{gridcode}

Theorem \ref{thm_1} guarantees the optimality in the union system. However, in practice, the optimal combination for an arbitrary areal unit may use operations other than union. Here, we further take subtraction into account (i.e., by subtracting the complementary area from a coarser grid), and attempt to find other feasible and better combinations to represent modifiable areal units. 

\vspace{-.5em}
\begin{figure}[h]
  \centering
  \includegraphics[width=0.99\linewidth]{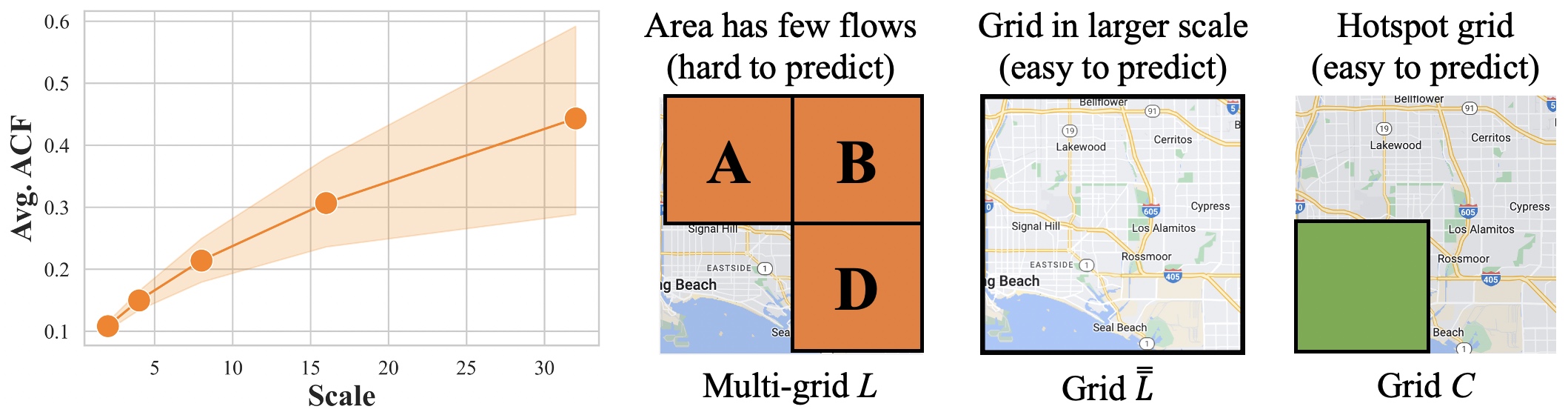}
  \caption{\textbf{Left}: Scale vs. predictability (with colored confidence intervals). \textbf{Right}: A multi-grid and two single grids with different predictability.}
  \vspace{-.7em}
  \label{fig: multigrid_motivation}
\end{figure}

For instance, as shown in Fig.~\ref{fig: multigrid_motivation}, if we take subtraction as a potential operation, there may be at least two ways to represent multi-grid $L$\footnote{A multi-grid consists of several adjacent grids on the same scale.}. The first is using three grids (i.e., A, B, and D) with union operations; another is using the coarse-scale grid $\Bar{\Bar{L}}$ to subtract the fine-scale grid $C$. In practice, the latter has certain probabilities to perform better than the former one, as coarse-scale grids may be easier to predict.

Specifically, as pointed out in previous research \cite{RegionGen_2023}, the Auto-Correlation Function (ACF) is a useful proxy for measuring regions' predictability. Our analysis confirms that areas with high flows generally exhibit better predictability, as indicated by larger ACF values (the left figure in Fig.~\ref{fig: multigrid_motivation}). We calculated the average ACF of each grid at different scales and found that coarser scales are generally easier to predict, with higher average ACF values.

Therefore, when querying a poor-predictability region where the upper grid and complementary area have better predictability, it may be advantageous to subtract the complementary area from a coarser grid. For instance, in Fig.~\ref{fig: multigrid_motivation}, Grid $C$ is the complementary area of multi-grid $L$ under $\Bar{\Bar{L}}$. Suppose multi-grid $L$ has few flows and poor predictability while grids $\Bar{\Bar{L}}$ (coarse-scale) and $C$ (hotspot) are easy to predict. In this case, subtracting Grid $\Bar{\Bar{L}}$ from Grid $C$ can result in improved estimates for multi-grid $L$.

\vspace{-0.8em}
\begin{figure}[h]
  \centering
  \includegraphics[width=0.8\linewidth]{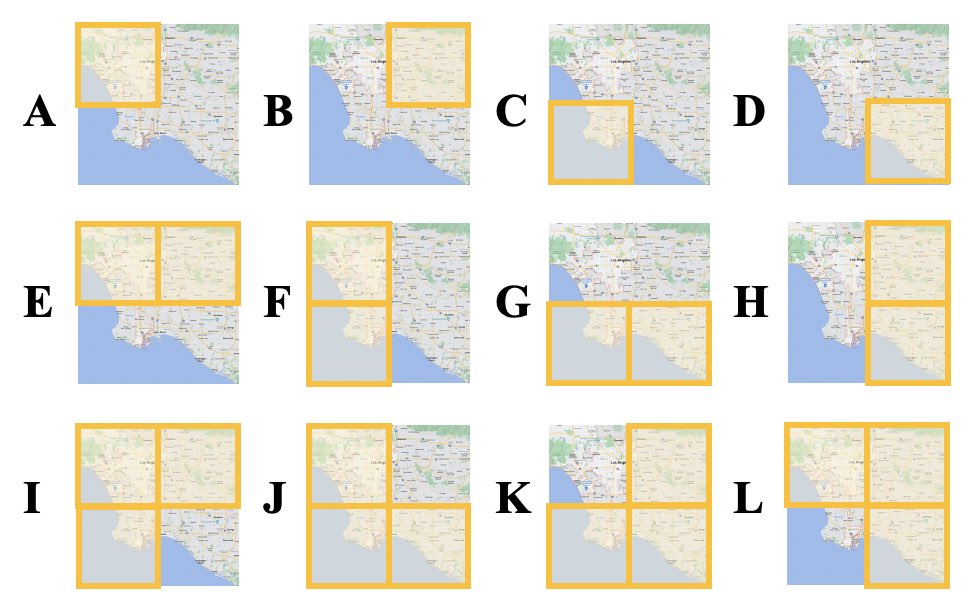}
  \vspace{-0.6em}
  \caption{Example of grid and multi-grid codes.}
  \vspace{-.7em}
  \label{fig: grid_code}
\end{figure}

To streamline the search process for further considering the subtraction operation, we introduce a grid coding rule for representing multi-grids. As illustrated in Fig.~\ref{fig: grid_code}, this involves using a merging window size of 2 and allowing each multi-grid to consist of up to three single grids. Grids $A$-$D$ represent four single grids whose optimal combinations were determined in Sec.~\ref{optim_search_union}, while Grids $E$-$H$ and $I$-$L$ are multi-grids comprising two and three single grids respectively. 

The search for multi-grids is conducted after getting optimal combinations for single grids. As a result, the search outcomes are at least as good as those of the single grid search that considers only union operations (see Theorem \ref{thm_2}). For example, the optimal combination of multi-grid $L$ in Fig.~\ref{fig: multigrid_motivation} is selected from the following two situations.
\begin{equation}
    \Lambda^*(L) \in \{\Lambda^*(A)+\Lambda^*(B)+\Lambda^*(D), \Lambda^*(\Bar{\Bar{L}})-\Lambda^*(C)\} \label{eq: multi_grid_L}
\end{equation}
where $\Bar{\Bar{L}}$ is the coarser gird containing multi-grid $L$ in the upper scale. During the search process, we will record the best combinations for all the single grids and multi-grids in Fig.~\ref{fig: grid_code}, which can then be directly applied to determine the optimal combinations for modifiable areal units.

\begin{thm} \label{thm_2}
The subtraction operations lead to solutions that are either better or equivalent to those obtained through the optimal combination search by union operations.
\end{thm}

\textit{Proof}. Let us consider the case in Fig. \ref{fig: multigrid_motivation}, which applies to other multi-grids as well. Eq. \ref{eq: multi_grid_L} shows that we obtain the optimal combination of $L$ by either uniting two single grids or subtracting the complementary area from a coarser grid. It makes the obtained solutions either be matched (i.e., $\Lambda^*(A) + \Lambda^*(B)$) or surpass (i.e., $\Lambda^*(\Bar{\Bar{L}})-\Lambda^*(C)$) the combination searched by union operations. $\blacksquare$

\subsubsection{Extended Quad-tree Index} \label{quadtree}

Once optimal combinations for each (multi-)grid are determined, they can be used for predicting region queries. Storing optimal combinations for single and multiple grids incurs a space complexity of $O(HW)$ each. Notably, $H$ and $W$ are often larger than 100, and rapid response times are crucial for downstream services in online prediction scenarios. Retrieving these optimal combinations in a linear table can be time-consuming and impractical.

\vspace{-.8em}
\begin{figure}[h]
  \centering
  \includegraphics[width=0.92\linewidth]{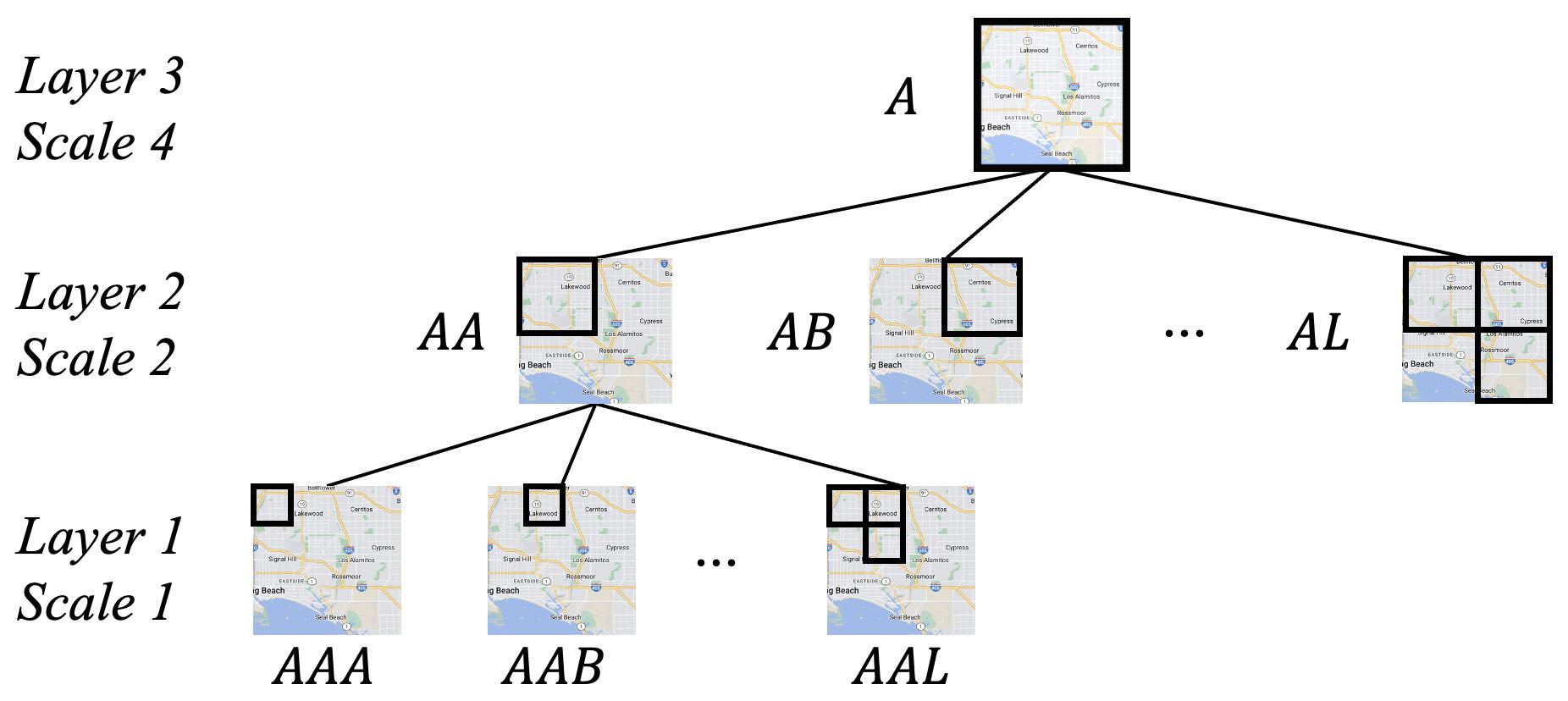}
  \vspace{-.5em}
  \caption{Example of an extended quad-tree with three layers.}
  \vspace{-.5em}
  \label{fig: quadtree}
\end{figure}

Therefore, we propose using a quad-tree to index the optimal combinations, as it is commonly used in spatial databases \cite{quad_r_tree_2002, Il_quadtree_2013}. However, since there may be more than four child grids for a coarse grid (e.g., a coarse grid may have twelve child grids, including four single grids and eight multi-grids as shown in Fig.~\ref{fig: grid_code}), we extend the basic quad-tree to allow nodes to have up to 12 child nodes. In Fig.~\ref{fig: quadtree}, we present an extended quad-tree example with three layers, where each grid in the hierarchical structure has a unique code. The introduction of the extended quad-tree reduces the time complexity of retrieving the optimal combination from $O(HW)$ to $O(\log(HW))$.

\subsection{Modifiable Areal Units Prediction} \label{decomposition}
\subsubsection{Hierarchical Decomposition}

To utilize the hierarchical structure and benefit from the found optimal combinations for grids, we first decompose arbitrary region queries into several hierarchical grids by Algorithm~\ref{alg: decomposition}. Fig.~\ref{fig: decomposition} gives an example of hierarchical region decomposition, which involves three scales of grids. The orange region query is decomposed into four hierarchical grids consisting of a coarsest grid, two medium-sized grids, and a multi-grid in the finest scale.

\subsubsection{Grid Indexing} 
After decomposing the region into hierarchical grids, we encode them into indexes based on the coding rule mentioned in Sec. \ref{gridcode}. Fig. \ref{fig: decomposition} gives an example of grid indexing. The orange region query is decomposed into four hierarchical grids, indexed as $B$, $AB$, $DB$, and $ADL$. These indexes allow us to quickly retrieve the optimal combination for the decomposed grids from the pre-constructed extended quad-tree. We use these optimal combinations to get the predictions of hierarchical grids (e.g., through union or subtraction operations) and sum them to obtain the predicted results of the region query.

\section{Experiments}
In this section, we evaluate the performance of \textit{One4All-ST} on two ST datasets to answer the following questions:

\textbf{RQ1}: Compared to the baseline methods (including fine-grained methods and multi-scale methods), how does our method perform in both effectiveness and efficiency?

\textbf{RQ2}: How does the optimal combination benefit the performance? 

\textbf{RQ3}: Compared with previous multi-scale ST networks \cite{STRN_2021, MC_STGCN_2022}, how do our proposed components (i.e., the hierarchical spatial modeling and scale normalization module) affect the performance? 

\textbf{RQ4}: How does the hierarchical structure (i.e., the merging window size) impact the performance? 


\subsection{Experimental Setup} \label{dataset_setup}

\subsubsection{Datasets} We collect two ST datasets including taxi trips and freight transport orders. We choose the last 20\% duration in each dataset as the test set, the 10\% data before the test for validation, and the remaining 70\% for training. We predict taxi/truck demand for the next hour. 

\textbf{Taxi NYC Dateset}. We collect the taxi trip dataset from NYC's open data portal\footnote{https://www.nyc.gov/site/tlc/about/tlc-trip-record-data.page}. The whole dataset spans over 10 years, and we use the records from Jan. 2013 to Mar. 2013, which amount to over 36,000,000 records. These records include pick-up/drop-off times, locations, and trip distances.

\textbf{Freight Transport Dataset}. The freight transport dataset is gathered from a world-leading online transportation company, including the freight transport orders within a metropolis from Oct. 2020 to Aug. 2021. The freight transport order typically takes place like this: users send orders online, and truck owners receive orders online and provide transportation services. This dataset contains over 7,000,000 records, with each order record containing the start time and location (longitude and latitude).

We partition the area of interest in both datasets into 128$\times$128 grids, each with a size of 150m$\times$150m, which is consistent with previous fine-grained prediction settings \cite{STRN_2021,overcome_ufi_2023}. The hierarchical structure $P=\{1,2,4,8,16,32\}$ is created by setting the maximum scale to 32 and using a merging window size of 2.

\subsubsection{Evaluation Metrics}
We exploit two widely used metrics including RMSE (Root Mean Square Error) and MAPE (Mean Absolute Percentage Error) to evaluate the performance of predictions \cite{ssl_traffic_2023,uncertainty_quant_2023,dynamic_hypergraph_2023}. 

\subsubsection{Prediction Tasks}
We choose four tasks with different scales to evaluate the capability for predicting arbitrary MAU in both datasets.
Task 1 predicts flows on hexagons or census tracts with an average spatial scale of 0.3 km$^2$, enabling detailed analysis tasks like fine-grained taxi flow prediction \cite{joint_demand_2021, zheng_deepstd_2020}.
Task 2 predicts flows on tertiary road map segments whose average spatial scale is 0.6 km$^2$. Such spatial scale is suitable for prediction tasks on function areas (e.g., residential quarter) \cite{wang_functional_2023}.
Task 3 predicts flows on secondary road map segments with an average spatial scale of 1.3 km$^2$, which is applicable for supply-demand prediction tasks \cite{geng2019spatiotemporal, ST_MetaNet_2022}.
Task 4 predicts flows on primary road map segments with an average spatial scale of 4.8 km$^2$. This scale aligns with typical community analysis tasks \cite{irregular_2022, sun_community_2016, flexible_partition_2020}.
Most region queries in the above tasks are defined by spatial semantic boundaries determined by visible features like streets and roads, enabling detailed analysis of demographics and socioeconomic factors \cite{flexible_partition_2020}. Task 1 of the Freight Transport dataset is the only exception, utilizing small hexagons (i.e., 350m$\times$350m) as fine-grained region queries. In transportation service applications like ride-sharing, fixed-shaped region queries such as hexagons are also prevalent \cite{CVNet2019}.
We obtain the polygon boundaries of the census tracts from NYC open data site\footnote{https://www.nyc.gov/site/planning/data-maps/open-data.page\#census}. We generate road map segments using the existing road segmentation method \cite{yuan2012segmentation}. The road network data are from OpenStreetMap\footnote{We download OSM data from http://download.geofabrik.de/}. In Fig~\ref{query_vis}, the upper and lower parts display the region queries for the Taxi NYC and Freight Transport datasets respectively.

\vspace{-.5em}
\newcommand{\subfigcol}{-3.2mm}
\begin{figure}[htbp]
\begin{minipage}[c]{1\linewidth}
\centering
\subfigure[\textit{Census tracts}]{
\includegraphics[width=0.24\linewidth]{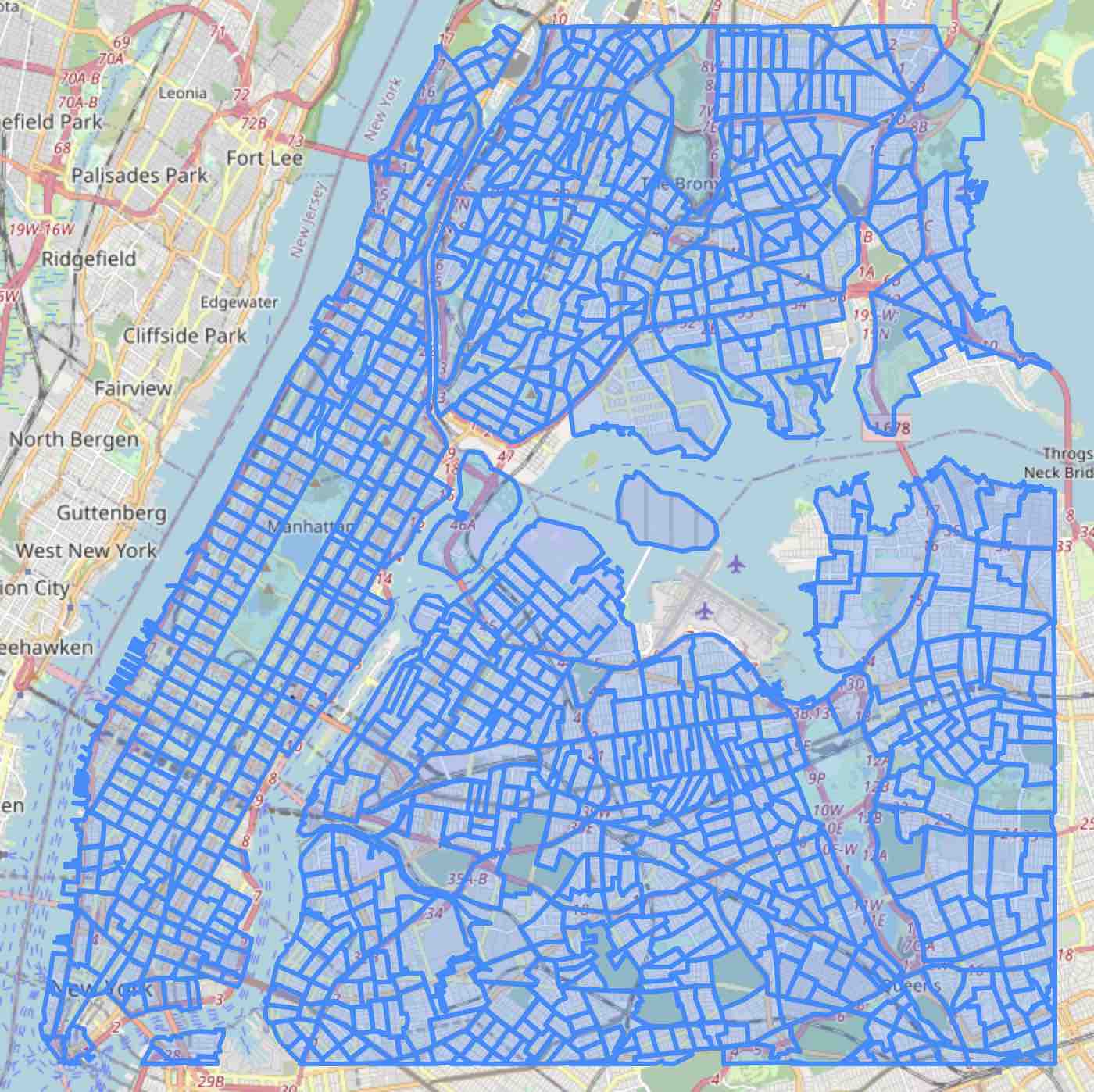}
    \label{nyc_task1}
}\hspace{\subfigcol}
\subfigure[\textit{Rd. Tertiary}]{
\includegraphics[width=0.24\linewidth]{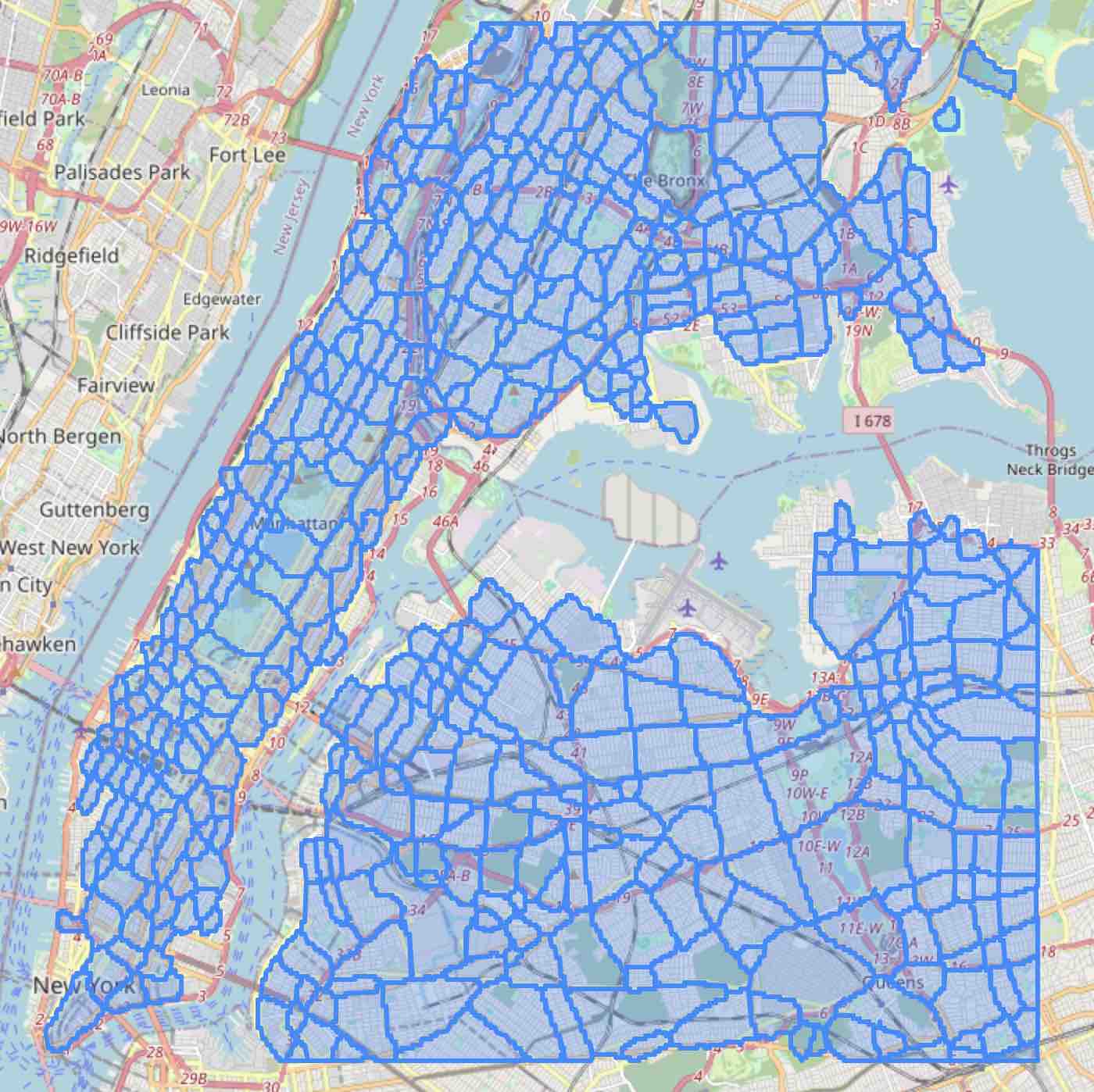}
    \label{nyc_task2}
}\hspace{\subfigcol}
\subfigure[\textit{Rd. Secondary}]{
\includegraphics[width=0.24\linewidth]{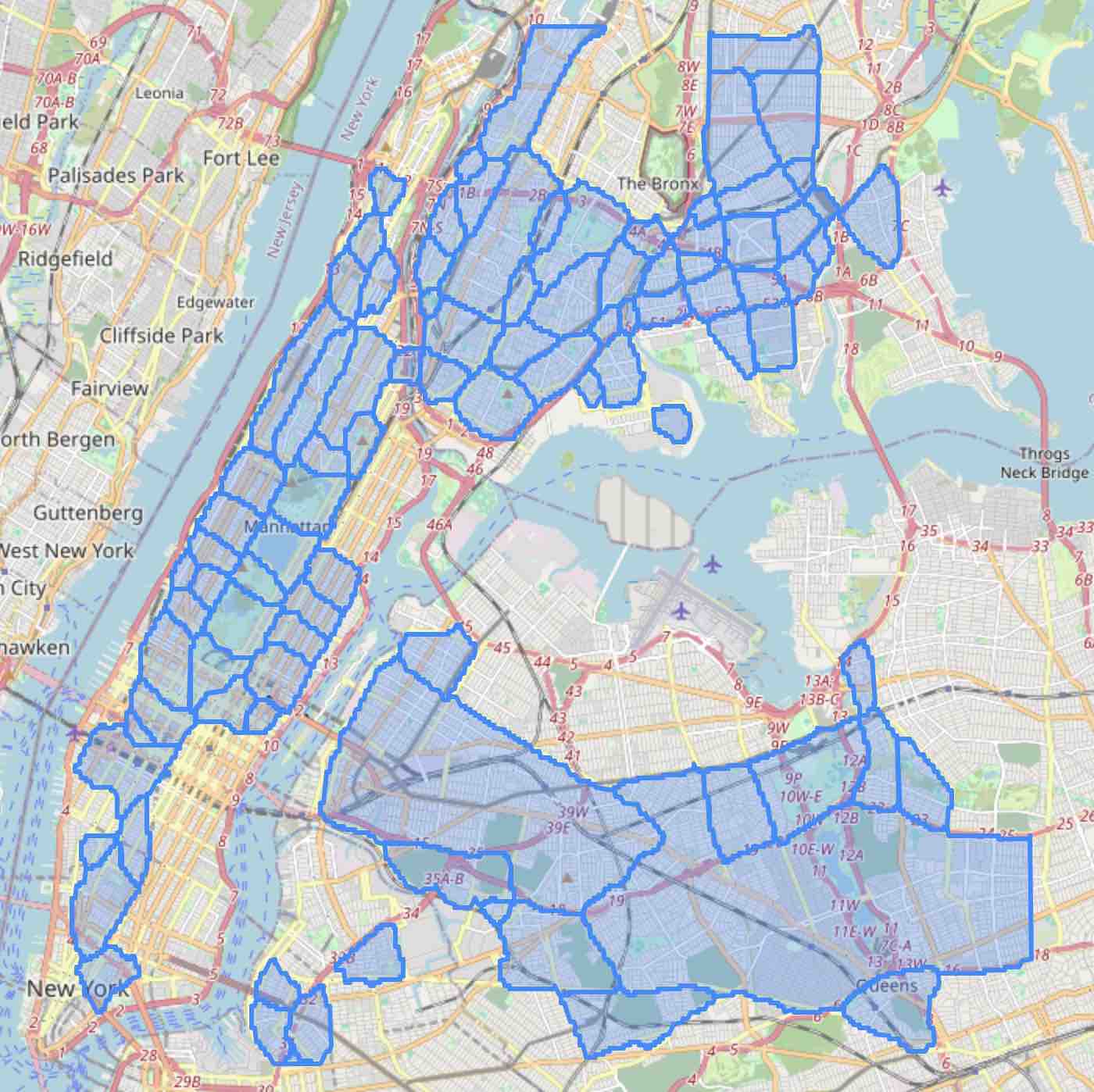}
    \label{nyc_task3}
}\hspace{\subfigcol}
\subfigure[\textit{Rd. Primary}]{
\includegraphics[width=0.24\linewidth]{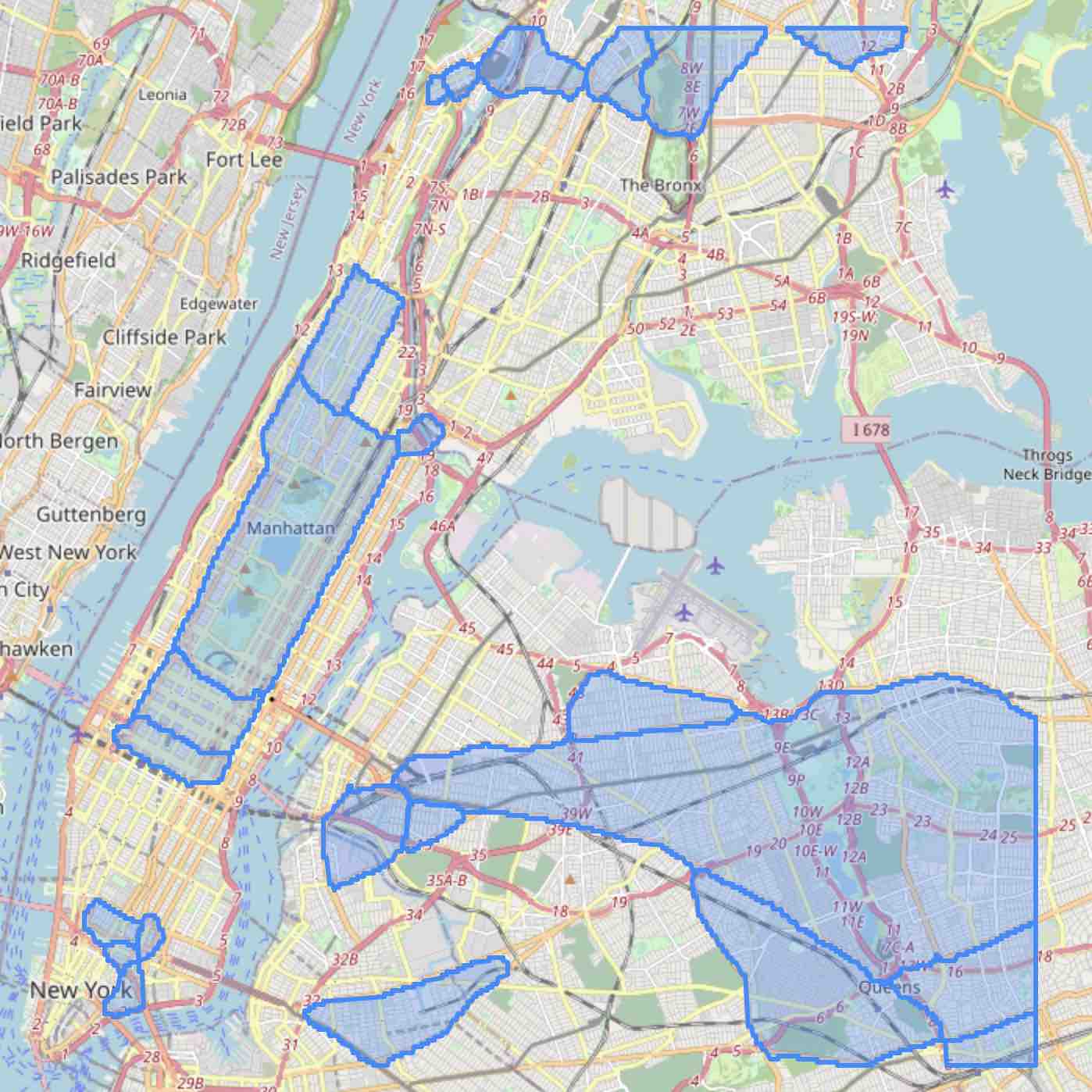}
    \label{nyc_task4}
}\hspace{\subfigcol}
\vspace{-0.5em}

\subfigure[\textit{Hexagons}]{
\includegraphics[width=0.24\linewidth]{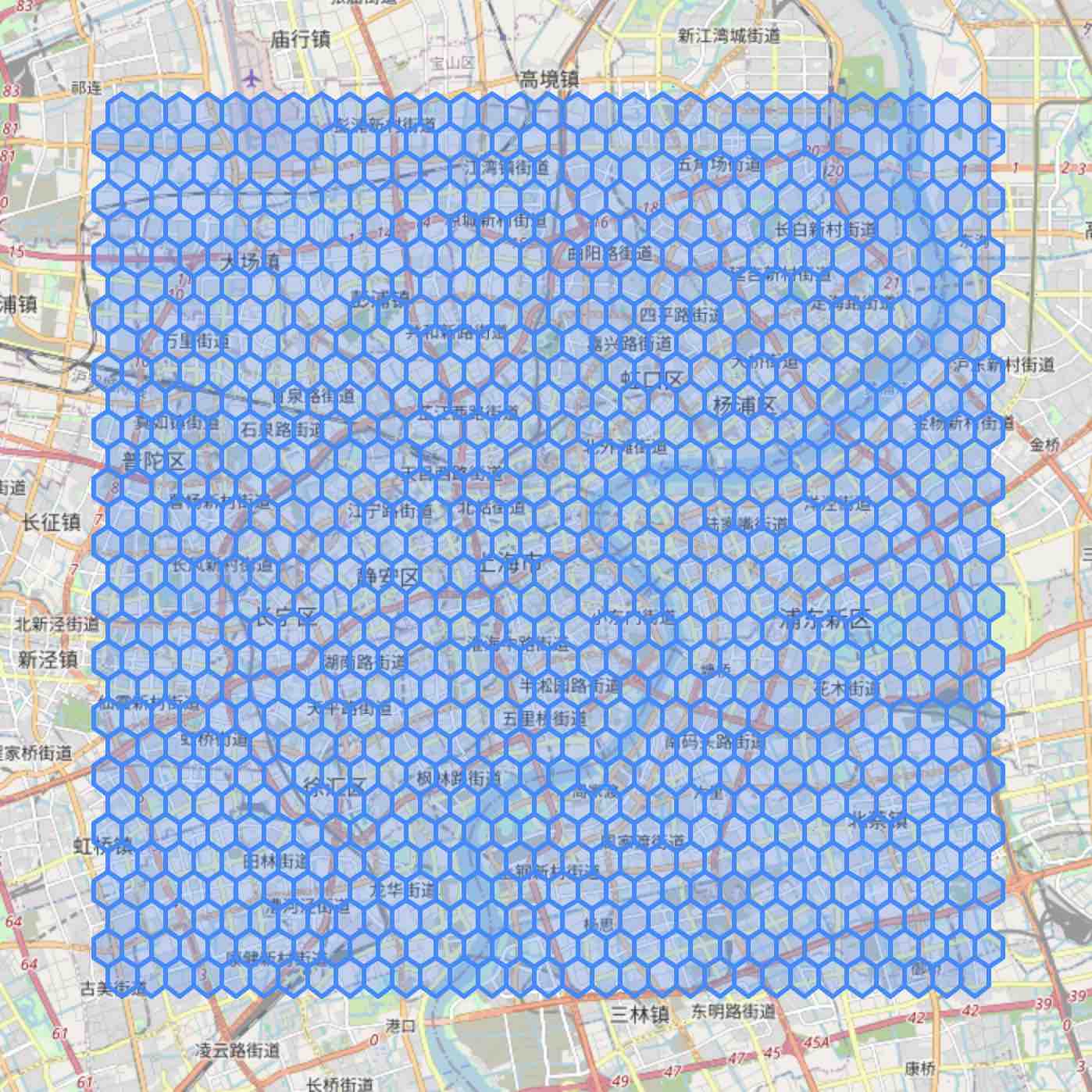}
    \label{sh_task1}
}\hspace{\subfigcol}
\subfigure[\textit{Rd. Tertiary}]{
\includegraphics[width=0.24\linewidth]{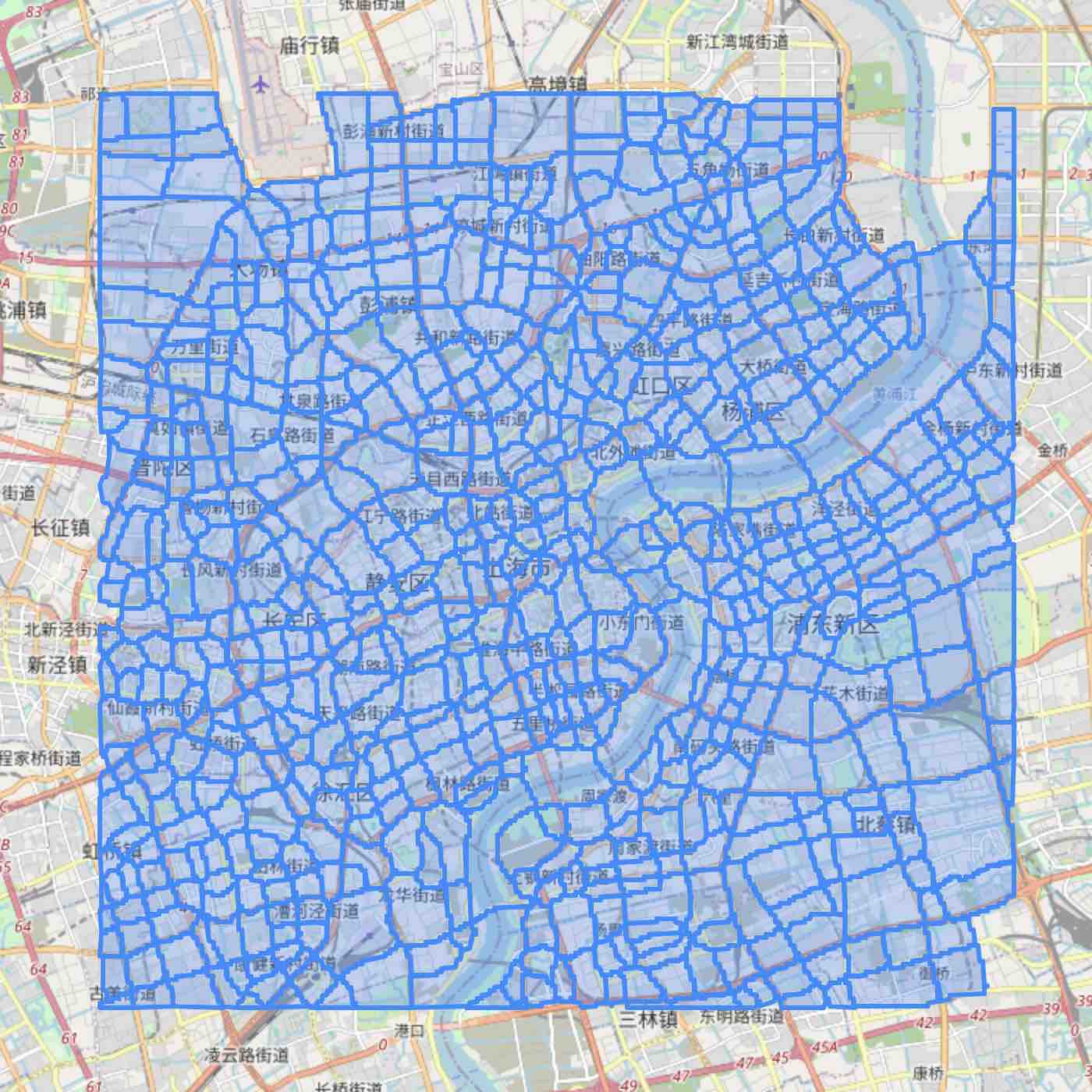}
    \label{sh_task2}
}\hspace{\subfigcol}
\subfigure[\textit{Rd. Secondary}]{
\includegraphics[width=0.24\linewidth]{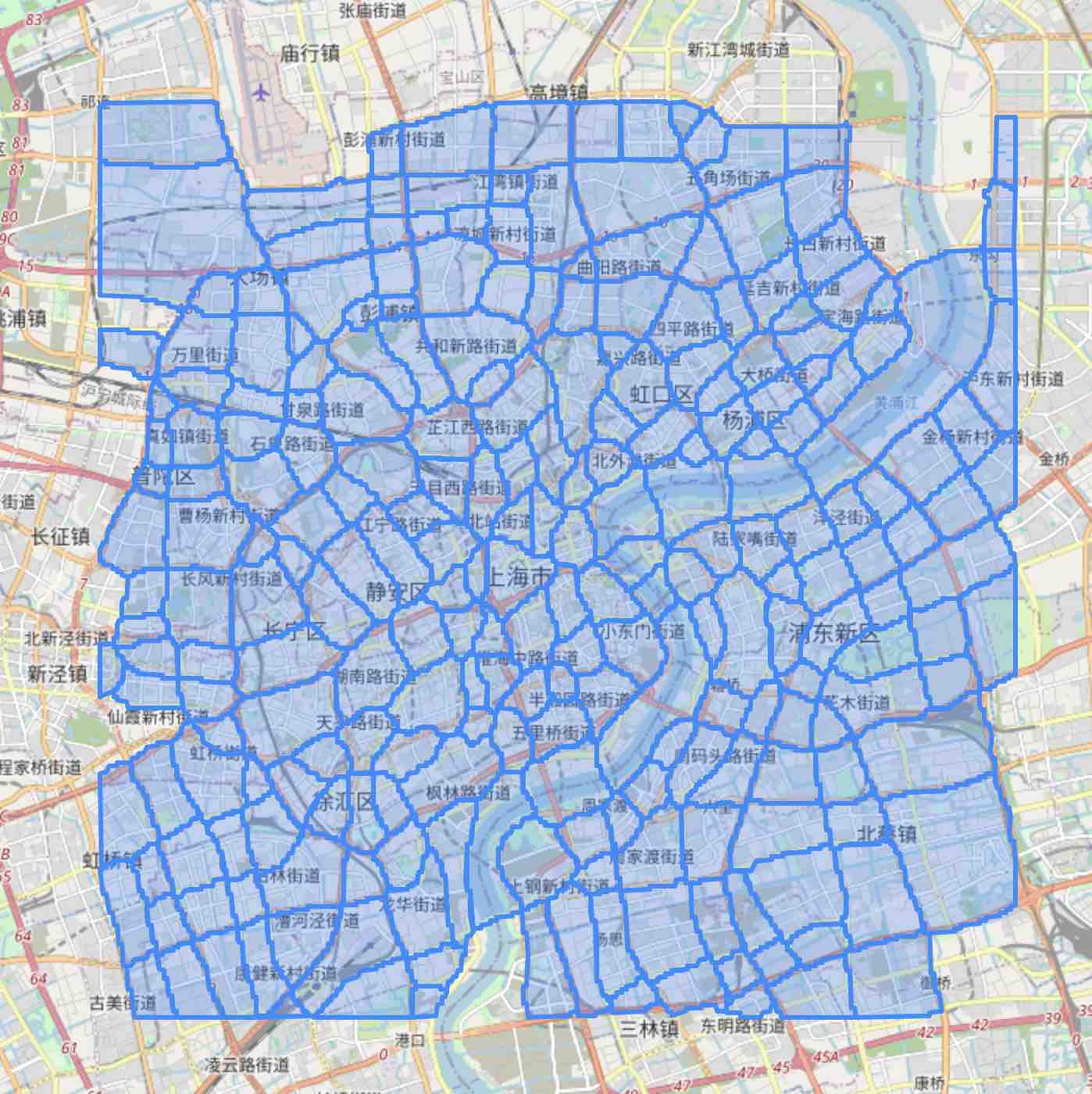}
    \label{sh_task3}
}\hspace{\subfigcol}
\subfigure[\textit{Rd. Primary}]{
\includegraphics[width=0.24\linewidth]{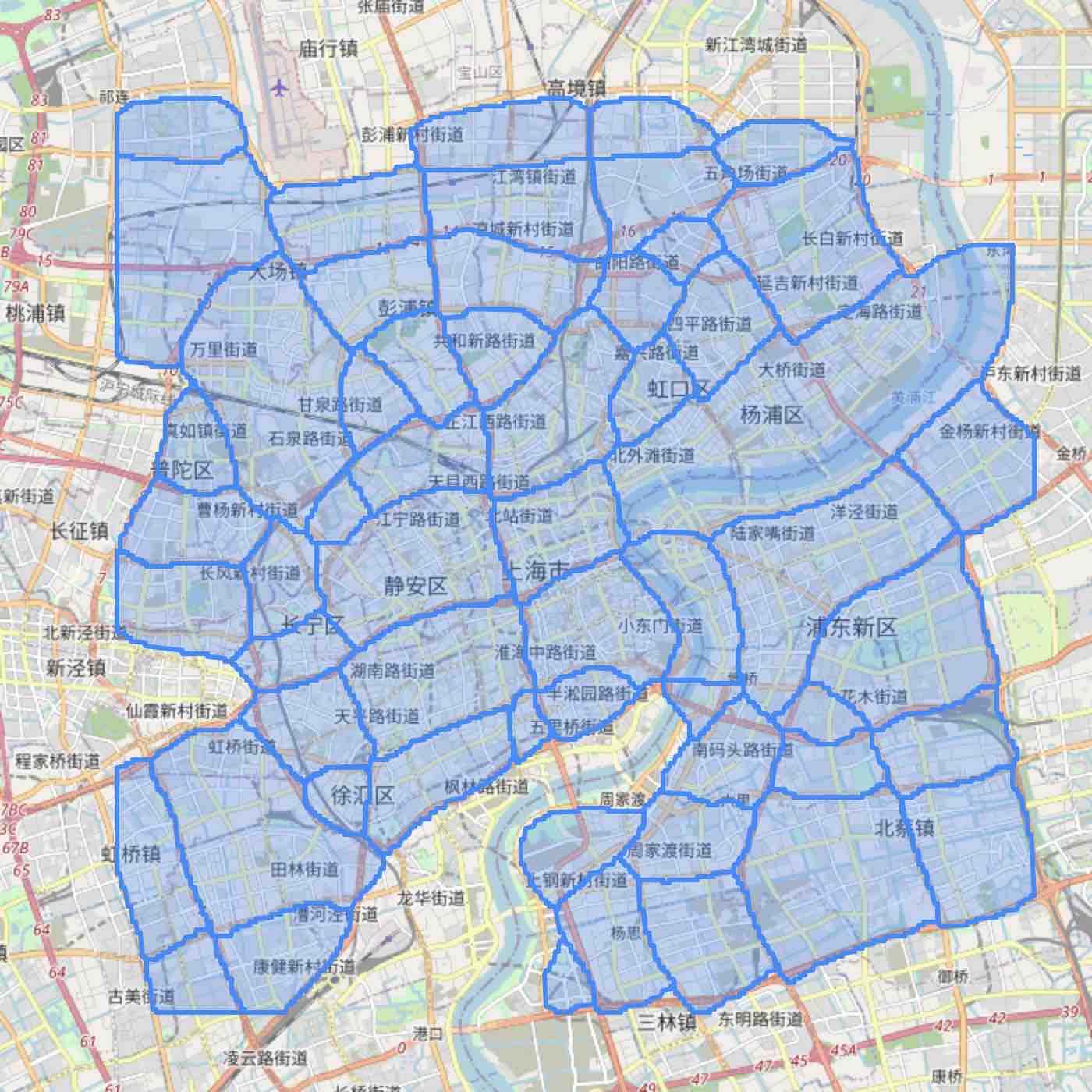}
    \label{sh_task4}
}\hspace{\subfigcol}
\vspace{-.5em}
\caption{Visualization of region queries. \textbf{Upper}: Census tracts (Task 1) and road map segments (Task 2, 3, 4) on Taxi NYC. \textbf{Lower}: Hexagons (Task 1) and road map segments (Task 2, 3, 4) on Freight Transport.}
\label{query_vis}
\end{minipage}
\vspace{-.5em}
\end{figure}

\begin{table*}[t]
\small
\centering
\caption{Results on the Taxi NYC and Freight Transport datasets. The best two results are highlighted (\textbf{\textit{best}} is in bold and \textit{italic}, \textbf{second-best} is in bold). \textit{M-ST-ResNet} and \textit{M-STRN} are our enhanced multi-scale models from \textit{ST-ResNet} and \textit{STRN}, respectively, which adopt our proposed optimal combination search module to improve performance on modifiable areal queries. }
\resizebox{1\textwidth}{!}{
\setlength{\tabcolsep}{1mm}{
\begin{tabular}{lcccccccccccccccccccccccccccccccccc}
\toprule
\multirow{3}{*}{} & \multicolumn{8}{c}{\textbf{Taxi NYC}} & \multicolumn{8}{c}{\textbf{Freight Transport}} \\
\cmidrule(lr){2-9} \cmidrule(lr){10-17}
& \multicolumn{2}{c}{\textit{Task 1}} & \multicolumn{2}{c}{\textit{Task 2}} & \multicolumn{2}{c}{\textit{Task 3}} & \multicolumn{2}{c}{\textit{Task 4}} & \multicolumn{2}{c}{\textit{Task 1}} & \multicolumn{2}{c}{\textit{Task 2}} & \multicolumn{2}{c}{\textit{Task 3}} & \multicolumn{2}{c}{\textit{Task 4}}\\
\cmidrule(lr){2-3} \cmidrule(lr){4-5} \cmidrule(lr){6-7} \cmidrule(lr){8-9} \cmidrule(lr){10-11} \cmidrule(lr){12-13} \cmidrule(lr){14-15} \cmidrule(lr){16-17}
& RMSE & MAPE & RMSE & MAPE & RMSE & MAPE & RMSE & MAPE & RMSE & MAPE & RMSE & MAPE & RMSE & MAPE & RMSE & MAPE \\
\midrule
\multicolumn{2}{l}{\textbf{Baselines}} \\
\textit{HM} & 21.95 & 0.130 & 29.52 & 0.122 & 60.50 & 0.124 & 138.9 & 0.130 & 1.745 & 0.370 & 1.928 & 0.384 & 2.374 & 0.387 & 4.390 & 0.313 \\
\textit{XGBoost} & 19.09 & 0.116 & 25.40 & 0.111 & 53.60 & 0.115 & 137.3 & 0.110 & 1.788 & 0.347 & 1.982 & 0.371 & 2.421 & 0.390 & 4.370 & 0.325\\
\textit{ST-ResNet} & 19.14 & 0.117 & 24.80 & 0.108 & 49.85 & 0.109 & 126.6 & 0.100 & 1.684 & 0.336 & 1.914 & 0.361 & 2.333 & 0.369 & 4.047 & 0.295\\
\textit{GWN} & 18.80 & 0.125 & 24.55 & 0.105 & 49.72 & 0.104 & 117.5 & 0.098 & 1.693 & 0.337 & 1.879 & 0.351& 2.262 & 0.356 & 3.991 & 0.292 \\
\textit{ST-MGCN} & 19.05 & 0.118 & 25.47 & 0.109 & 50.81 & 0.110 & 126.2 & 0.098 & 1.765 & 0.346 & 1.963 & 0.378 & 2.417 & 0.399 & 4.411 & 0.361 \\
\textit{GMAN} & 18.86 & 0.124 & 25.16 & 0.107 & 50.80 & 0.103 & 123.6 & 0.096 & 1.721 & 0.360 & 1.891& 0.362 & 2.304 & 0.375 & 4.100 & 0.304 \\
\textit{STRN} & 18.68 & 0.111 & 24.92 & 0.109 & 51.93 & 0.114 & 131.6 & 0.104 & 1.653 & 0.333 & 1.917 & 0.363 & 2.343 & 0.380 & 4.112 & 0.312 \\
\textit{MC-STGCN} & 19.19 & 0.119 & 25.58 & 0.111 & 51.76 & 0.113 & 126.3 & 0.105 & 1.758 & 0.370 & 1.945 & 0.384 & 2.397 & 0.396 & 4.412 & 0.330 \\
\textit{STMeta} & 19.04 & 0.109 & 25.99 & 0.114 & 53.26 & 0.122 & 134.4 & 0.103 & 1.726 & \textbf{0.332} & 1.900 & 0.356 & 2.308 & 0.371 & 4.023 & 0.322 \\
\midrule
\multicolumn{4}{l}{\textbf{Our Enhanced Methods}} \\
\textit{M-ST-ResNet} & \textbf{18.14} & \textbf{0.108} & \textbf{23.58} & \textbf{0.103} & \textbf{46.21} & \textbf{0.102} & \textbf{\textit{109.9}} & \textbf{0.083} & 1.683 & 0.336 & 1.856 & 0.344 & 2.241 & 0.350 & \textbf{\textit{3.769}} & \textbf{0.275}\\
\textit{M-STRN} & 18.65 & 0.110 & 24.67 & 0.107 & 49.28 & 0.107 & 121.8 & 0.093 & \textbf{1.652} & \textbf{0.332} & \textbf{1.842} & \textbf{0.341} & \textbf{2.226} & \textbf{0.340} & 3.846 & \textbf{\textit{0.271}}\\
\midrule
\textit{One4All-ST} & \textbf{\textit{17.48}} &  \textit{\textbf{0.104}} & \textbf{\textit{22.74}} & \textbf{\textit{0.099}} & \textbf{\textit{44.45}} & \textbf{\textit{0.099}} & \textbf{110.2} & \textit{\textbf{0.082}} & \textbf{\textit{1.649}} & \textbf{\textit{0.330}} & \textbf{\textit{1.798}} & \textbf{\textit{0.331}} & \textbf{\textit{2.181}} & \textbf{\textit{0.336}} & \textbf{3.778} & \textbf{0.275} \\
\bottomrule
\end{tabular}}}
\vspace{-1em}
\label{table: main_results}
\end{table*}

\subsubsection{Baselines and Enhanced Methods}
We compare \textit{One4All-ST} with the following baselines. Most baselines, except \textit{MC-STGCN}~\cite{MC_STGCN_2022}, are single-scale ST models that predict at the finest atomic scale. These single-scale methods aggregate results on atomic grids to predict region queries. In contrast, \textit{MC-STGCN}, a bi-scale baseline, simultaneously predicts at both the finest atomic and coarse-grained cluster scale. The clustering process takes in geographic proximity information and historical crowd flow as described in the original paper \cite{MC_STGCN_2022}. \textit{MC-STGCN} aims to utilize cluster predictions whenever possible since coarse scales are generally easier to predict as shown in the left chart in Fig.~\ref{fig: multigrid_motivation}. Specifically, for each region query, \textit{MC-STGCN} uses cluster predictions if the clusters fall within the region query area, with the remaining complementary area predicted at the finest atomic scale.
\begin{itemize}[leftmargin=0.4cm]
    \item \textbf{\textit{HM}} (History Mean) predicts future demands using the mean value of the historical records.
    \item \textbf{\textit{XGBoost}} \cite{xgboost_2016} is a tree model that takes historical traffic data as features.
    \item \textbf{\textit{ST-ResNet}} \cite{zhang2017deep} applies residual convolution networks to capture spatial correlations.
    \item \textit{\textbf{GWN}} \cite{graph_wavenet_2019} (GraphWaveNet) proposes a data-driven method for adaptively learning spatial correlations.
    \item \textit{\textbf{ST-MGCN}} \cite{geng2019spatiotemporal} applies graph convolutions to capture multiple spatial relations.
    \item \textit{\textbf{GMAN}} \cite{GMAN_2020} proposes spatial and temporal attention mechanisms to capture the spatial and temporal correlations.
    \item \textbf{\textit{STRN}} \cite{STRN_2021} builds a coarse-grained scale that learns global spatial dependencies to enhance fine-grained predictions.
    \item \textbf{\textit{MC-STGCN}} \cite{MC_STGCN_2022} designs a cross-scale spatial-temporal feature learning module and gives bi-scale ST prediction.
    \item \textbf{\textit{STMeta}} \cite{STMeta} captures multiple kinds of temporal correlations as well as heterogeneous spatial correlations. 
\end{itemize}
Moreover, we also implement two multi-scale models (i.e., \textit{M-ST-ResNet} and \textit{M-STRN}) by extending existing models. We train several ST models to generate predictions on the same scales as \textit{One4All-ST}. To fully use multi-scale predictions for modifiable areal units, we applied the proposed optimal combinations (Sec.~\ref{optimal_comb}) for these methods.

\subsubsection{Implementation Details} \label{implementation_details}
For fair comparisons, all methods use the same temporal inputs as \textit{One4All-ST}. These inputs consist of 17 historical observations, six closeness records, seven daily records, and four weekly records. The only exception is \textit{HM}, which uses one closeness record, three daily records, and one weekly record through grid search. Our experiment platform is a server with 8 CPU cores (Intel Core i9-9900K @ 3.60GHz), 16 GB RAM, and one GPU (NVIDIA GeForce RTX 2080). We use Python 3.6.5 with TensorFlow (1.13.1) in Ubuntu Linux 5.19.0-43-generic release.

\begin{table}[htbp]
\small
\centering
\caption{The computation cost comparison of deep models. We report the total computational costs of six models in \textit{M-ST-ResNet} and \textit{M-STRN}.}
\setlength{\tabcolsep}{3.3mm}{
\begin{tabular}{lccc}
\toprule
\multirow{2}{*}{Taxi NYC} & Training & Inference & \multirow{2}{*}{\# Parameters} \\
& (sec/epoch) & (sec) & \\
\midrule
\textit{ST-ResNet} & 21.35 & 4.41 & 0.59M\\
\textit{GWN} & 11.98   & 0.99 & 0.92M  \\
\textit{ST-MGCN} & 20.52  & 5.37 & 2.51M  \\
\textit{GMAN} & 34.12 & 0.90 & 0.22M \\
\textit{STRN} & 22.73 & 2.33  & 0.88M \\
\textit{MC-STGCN} & 12.17 & 2.68 & 1.68M\\
\textit{STMeta} & {20.42} & 4.15 & 1.25M \\
\midrule
\textit{M-ST-ResNet} & 47.00 & 8.88 & 0.59M$\times$6 \\
\textit{M-STRN} & 55.00 & 3.47 & 0.88M$\times$6 \\
\midrule
\textit{One4All-ST} & 25.54 & 3.65  & 0.72M\\
\bottomrule
\end{tabular}}
\label{table: efficiency}
\vspace{-1.5em}
\end{table}

\subsection{Experimental Result}
\subsubsection{Main Result}
Here, we compare the effectiveness and efficiency of \textit{One4All-ST} with baselines (\textbf{RQ1}). Table \ref{table: main_results} presents RMSE and MAPE for various tasks across baselines\footnote{We also tested methods using MAE and found consistent results with RMSE and MAPE. Due to page limitations, we only report RMSE and MAPE.}, enhanced methods, and \textit{One4All-ST}. Additionally, Table \ref{table: efficiency} reports the computation cost of deep models. These two tables further reveal several insightful observations.

First, recall that from task 1 to task 4, the scale of region queries is getting coarser. In Table \ref{table: main_results}, we observed that in existing methods, deep learning methods (e.g., \textit{ST-ResNet} and \textit{STRN}) perform slightly better than statistical learning methods (e.g., \textit{XGBoost}) in task 1 where most queries are on fine scales. More importantly, when applied to coarser-scale tasks, the advantage of deep learning methods in predicting coarse-scale queries becomes more apparent.
    
Second, when comparing \textit{ST-ResNet}/\textit{STRN} (single-scale prediction models) with \textit{M-ST-ResNet}/\textit{M-STRN} (enhanced multi-scale prediction models), considerable performance improvements were observed, especially for coarse-scale tasks. It highlights the crucial role of coarse-scale predictions for coarse-scale region queries. Additionally, it also suggests that aggregating fine-scale predictions is insufficient for achieving precise coarse-scale prediction results.

Third, although \textit{MC-STGCN} (a bi-scale model) does not perform as well as other deep methods (e.g., \textit{ST-ResNet} and \textit{STRN}) on fine-scale tasks, thanks to the presence of coarser scales, it outperforms them slightly in coarse-scale tasks. It further confirms that coarse-scale predictions are vital for coarse-scale region queries. Note that \textit{MC-STGCN} uses separate spatial learning modules at different scales, resulting in much more trainable parameters (i.e., 1.68M) and increased cost compared to other methods.

Last, as shown in Table \ref{table: efficiency}, \textit{One4All-ST} is relatively lightweight in parameters (even with fewer parameters than \textit{STRN}, a single-scale model). However, it leads comprehensively in accuracy for all tasks. Specifically, on task 3, our method can achieve up to 10.6\% improvement over the best baseline in terms of RMSE on the Taxi NYC dataset. On the other side, compared with the enhanced methods (training separately without cross-scale information interaction), our approach achieves better results on most tasks, which confirms the effectiveness of our hierarchical structure and cross-scale modeling module. More importantly, our approach achieves better or comparable performance as the enhanced methods while using only 20\% parameters (Table \ref{table: efficiency}), which demonstrates the efficiency of our hierarchical ST network.

\subsubsection{Analysis of Optimal Combination Search}
To explore how the optimal combination search component improves performance (\textbf{RQ2}), we analyze queries whose optimal combinations are achieved via union or subtraction operations. There are three strategies for predicting arbitrary modifiable areal units: \textit{Direct}, \textit{Union}, and \textit{Union \& Subtraction}. \textit{Direct} predicts region queries by directly summing decomposed hierarchical grid predictions (by Algorithm \ref{alg: decomposition}) without considering optimal combination search. \textit{Union} applies optimal grid combinations obtained through union operations while \textit{Union \& Subtraction} considers both union and subtraction operations.

\newcommand{\threefigcol}{1mm}
\begin{figure*}[htbp]
\centering
\begin{minipage}[c]{0.24\linewidth}
\centering
\includegraphics[width=1\linewidth]{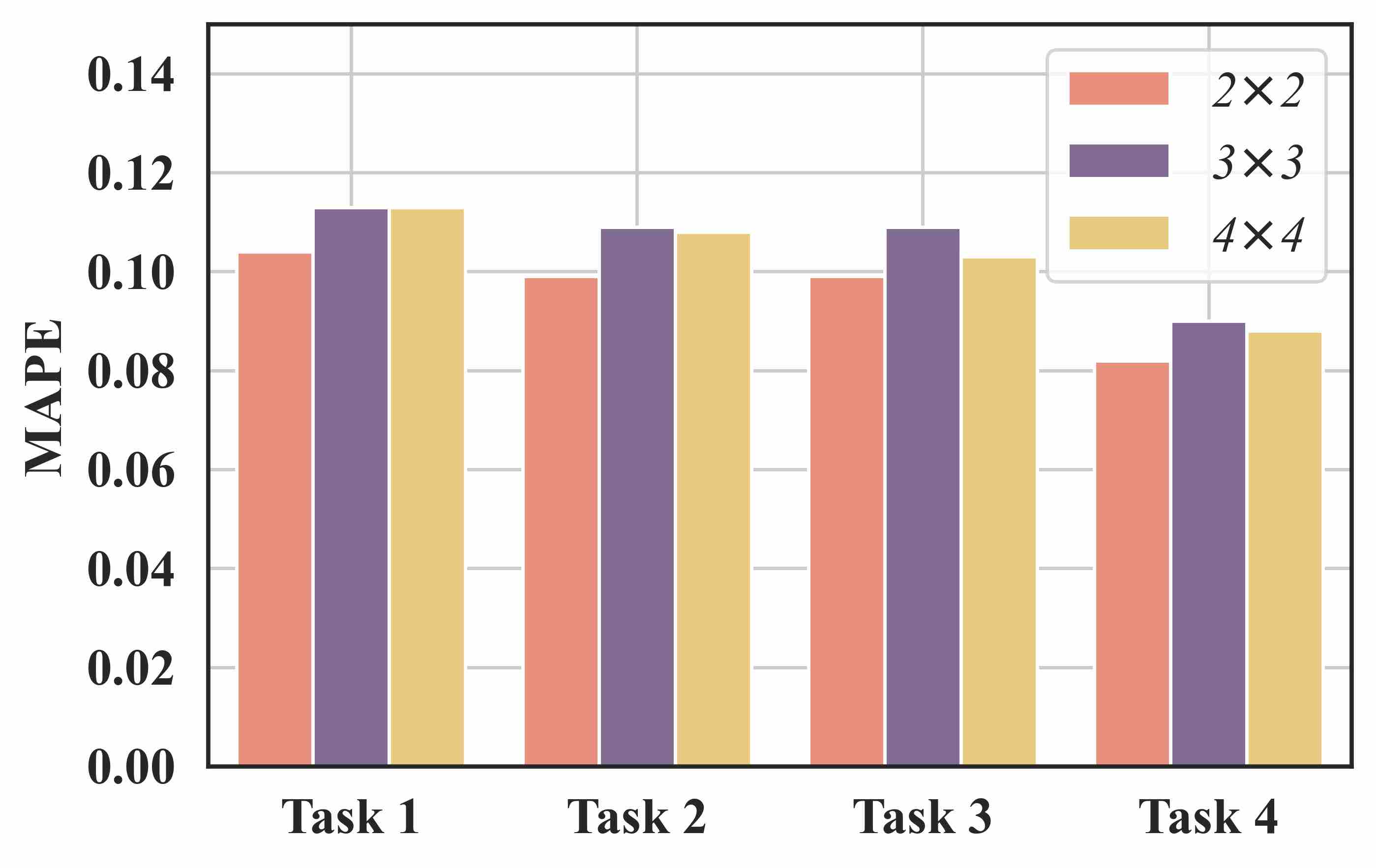}
\caption{Effect of hierarchical structure.}
\label{fig: mws_analysis}
\end{minipage}\hspace{\threefigcol}
\begin{minipage}[c]{0.24\linewidth}
\centering
\includegraphics[width=1\linewidth]{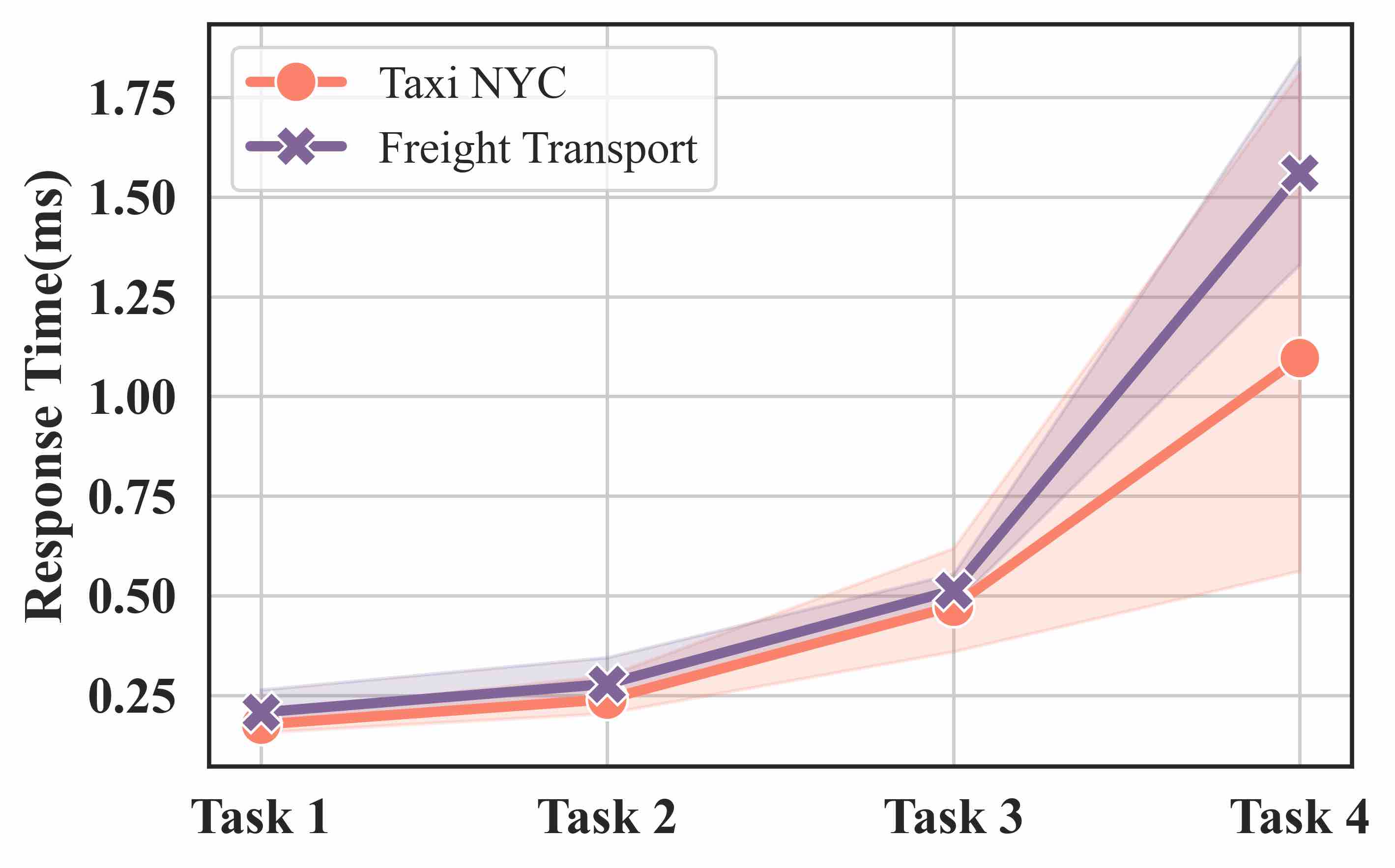}
\caption{Response time to region queries.}
\label{fig: response_time}
\end{minipage}\hspace{\threefigcol}
\begin{minipage}[c]{0.24\linewidth}
\centering
\includegraphics[width=1\linewidth]
{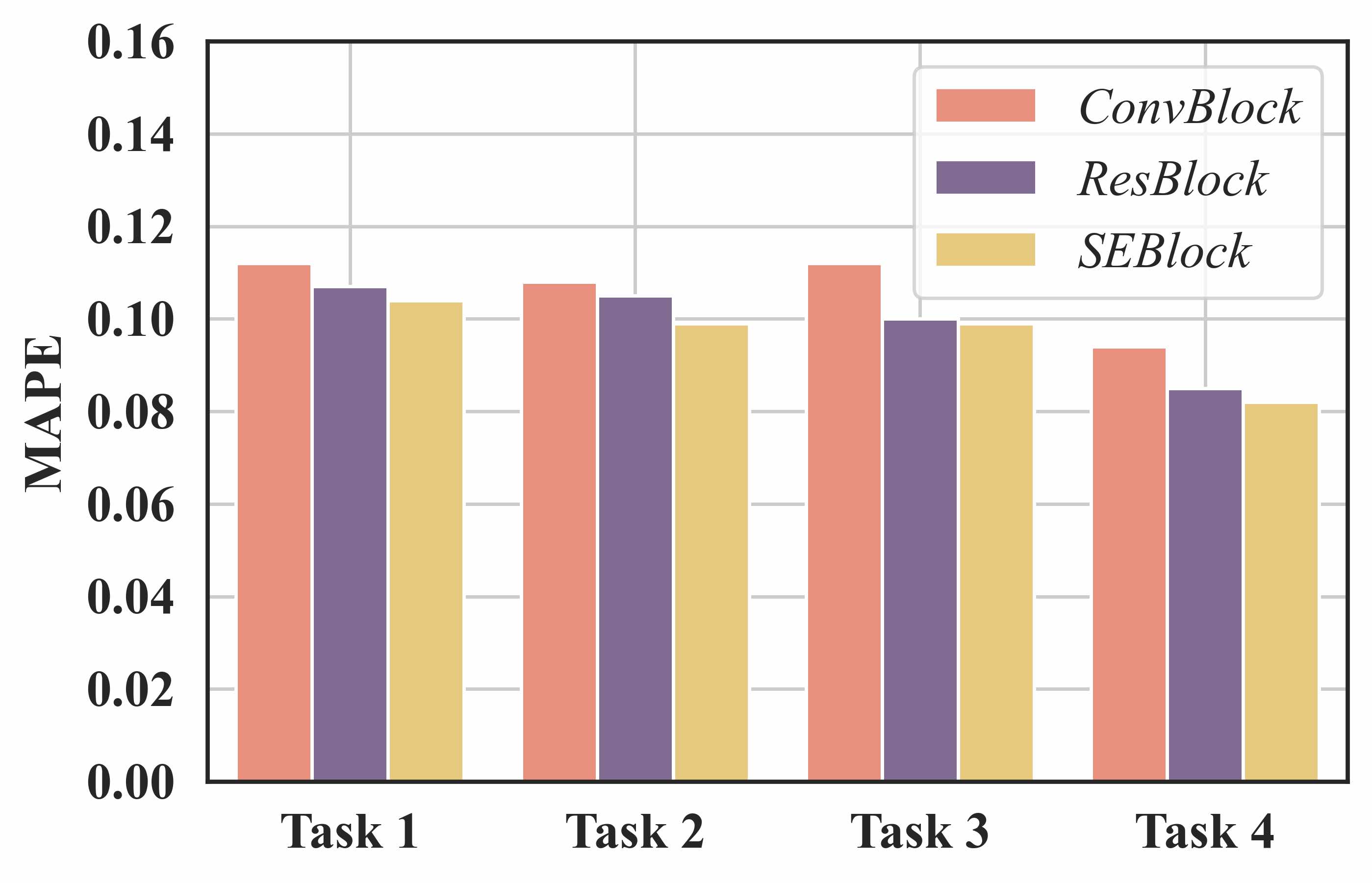}
\caption{Effect of spatial modeling block.}
\label{fig: effect_spatial_model_block}
\end{minipage}\hspace{\threefigcol}
\begin{minipage}[c]{0.24\linewidth}
\centering
\includegraphics[width=1\linewidth]
{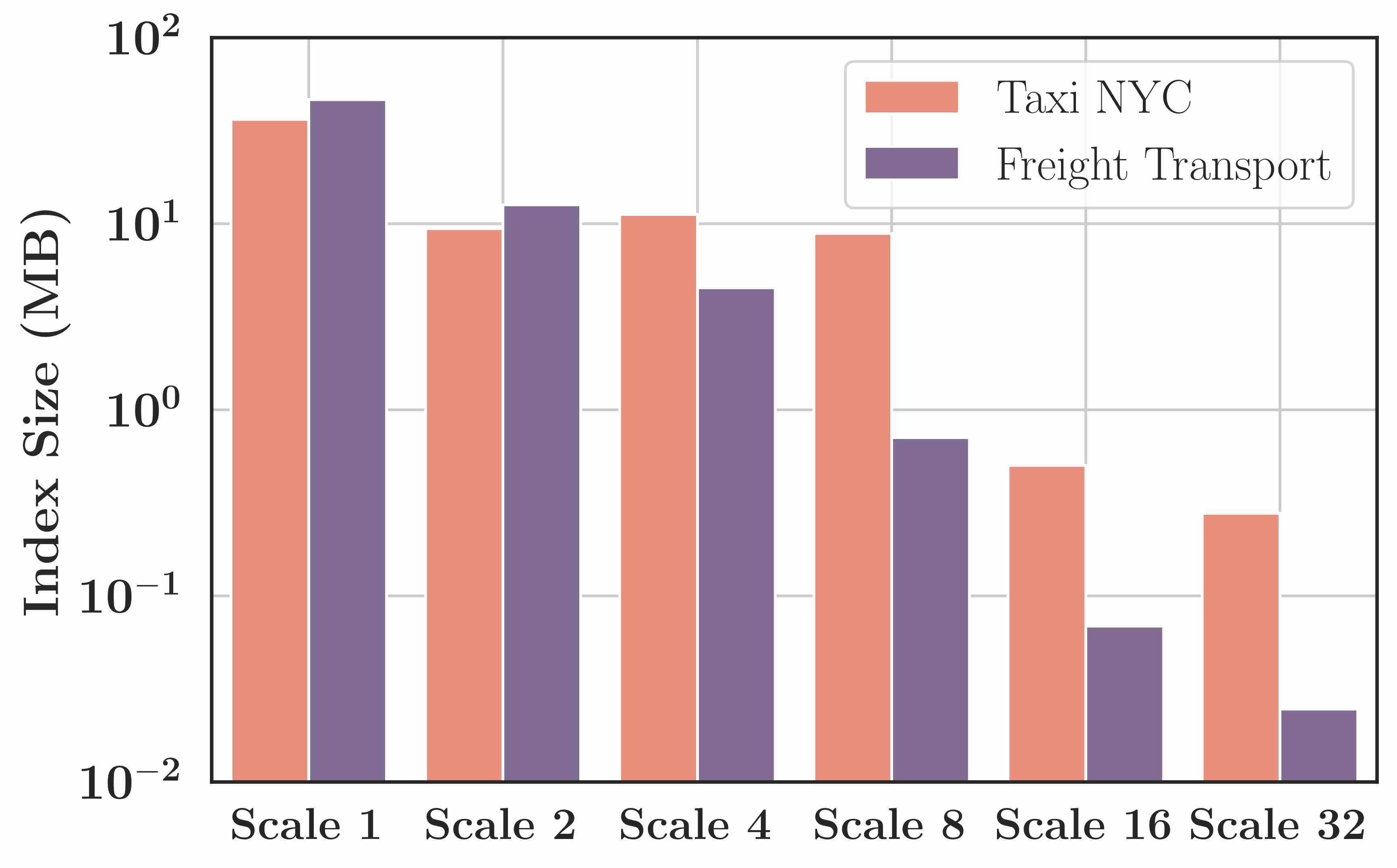}
\caption{Analysis of index size.}
\label{fig: index_size}
\end{minipage}
\vspace{-1em}
\end{figure*}

Table \ref{table: optimal_comb} lists the RMSE for these three strategies over four prediction tasks. 
In general, we observe more noticeable enhancements in employing optimal search with union/subtraction, particularly for coarse-scale tasks (Task 4). The possible reason may be that fine-scale tasks already exhibit optimal performance through the direct decomposition in Algorithm \ref{alg: decomposition} (i.e., utilizing the largest grids smaller than the query area for prediction); hence, the optimal search with union/subtraction cannot find better decomposition results. To verify this, we conduct an analysis of the queries leading to different grid decomposition for \textit{Union}/\textit{Union \& Subtraction} in comparison to \textit{Direct}. As anticipated, for fine-scale tasks like Task 1, only a marginal 7.14\%/8.14\% of queries show distinct decomposition results between \textit{Union}/\textit{Union \& Subtraction} and \textit{Direct}. Nevertheless, focusing on these queries with differing decomposition, we see a substantial 8.0\%/9.2\% improvement in Task 4 prediction accuracy. This underscores the value of seeking a superior decomposition through union/subtraction for performance enhancement.

Moreover, a comparison between \textit{Union \& Subtraction} and \textit{Union} demonstrates an increase of improved queries, rising from 11.8\% to 16.5\% for Task 3. This indicates that the introduction of subtraction operations can indeed yield better region query decomposition results in practical applications. Notably, the advantage of utilizing \textit{Union}/\textit{Union \& Subtraction} for optimal decomposition search also lies in its offline nature, thereby incurring no overhead for online prediction services.

\begin{table}[t]
\small
\centering
\caption{Results of three region query decomposition strategies on the Taxi NYC dataset. \textit{RMSE} is all the queries' average error; \textit{Prop.} (\%) is queries' proportion of different grid decompositions for \textit{Union}/\textit{Union \& Subtraction} in comparison to \textit{Direct}; \textit{Imprv.} (\%) is the prediction accuracy improvement of \textit{Union}/\textit{Union \& Subtraction} compared to \textit{Direct} specifically on these differently decomposed queries.}
\resizebox{0.485\textwidth}{!}{
\setlength{\tabcolsep}{1.5mm}{
\begin{tabular}{lccccccccccc}
\toprule 
 & \textbf{\textit{Direct}} & \multicolumn{3}{c}{\textbf{\textit{Union}}} & \multicolumn{3}{c}{\textbf{\textit{Union \& Subtraction}}} \\
\cmidrule(lr){2-2} \cmidrule(lr){3-5} \cmidrule(lr){6-8} 
& RMSE & \textit{Prop.} & \textit{Imprv.} & RMSE & \textit{Prop.} & \textit{Imprv.} & RMSE \\
\midrule
Task 1 & 17.53 & 7.16\% & 1.2\% & 17.51 & 8.14\% & 2.0\% & 17.48 \\
Task 2 & 23.02 & 10.1\% & 3.5\% & 22.75 & 12.9\% & 5.5\% & 22.74 \\
Task 3 & 45.41 & 11.8\% & 5.8\% & 44.62 & 16.5\% & 7.1\% & 44.45 \\
Task 4 &  113.8 & 11.6\% & 8.0\% & 110.6 & 12.1\% & 9.2\% & 110.2 \\
\bottomrule
\end{tabular}}}
\label{table: optimal_comb}
\vspace{-1.5em}
\end{table}

\subsubsection{Ablation Study for the Hierarchical Multi-scale Network}
We conduct an ablation study to verify the effectiveness of our proposed hierarchical spatial modeling and scale normalization modules for multi-scale ST learning (\textbf{RQ3}). \textit{One4All-ST (w/o HSM)} removes the hierarchical spatial modeling module, making every scale learn its ST representation from scratch instead of learning from the previous scale's representation. \textit{One4All-ST (w/o SN)} applies a single standard normalization transformation to all scales.

\begin{table}[t]
\small
\centering
\caption{Ablation results of the hierarchical multi-scale network. \textbf{\textit{best}} is in bold and \textit{italic}. `HSM' and `SN' are the abbreviations of Hierarchical Spatial Modeling and Scale Normalization.}
\begin{tabular}{cccccc}
\toprule
\multicolumn{2}{c}{\multirow{2}{*}{Taxi NYC}} & \textit{One4All-ST} & \textit{One4All-ST} & \multirow{2}{*}{\textit{One4All-ST}} \\
\multicolumn{2}{c}{} & (\textit{w/o HSM}) & (\textit{w/o SN}) & \\
\midrule
\multirow{2}{*}{Task 1} & RMSE & 18.36 & 34.59 & \textbf{\textit{17.48}} \\
& MAPE & 0.108 & 0.228 & \textbf{\textit{0.104}} \\
\midrule
\multirow{2}{*}{Task 2} & RMSE & 24.41 & 41.16 & \textbf{\textit{22.74}} \\
& MAPE & 0.107 & 0.184 & \textbf{\textit{0.099}} \\
\midrule
\multirow{2}{*}{Task 3} & RMSE & 49.14 & 69.46 & \textbf{\textit{44.45}} \\
& MAPE & 0.113 & 0.157 & \textbf{\textit{0.099}} \\
\midrule
\multirow{2}{*}{Task 4} & RMSE & 125.0 & 135.1 & \textbf{\textit{110.2}} \\
& MAPE & 0.091 & 0.150 & \textbf{\textit{0.082}} \\
\bottomrule
\end{tabular}
\label{table: ablation}
\vspace{-1.5em}
\end{table}

Table \ref{table: ablation} displays ablation results on the Taxi NYC dataset. Notably, \textit{One4All-ST (w/o HSM)} underperforms \textit{One4All-ST} at every scale, with the advantage of \textit{One4All-ST} becoming more pronounced as the scale increases (11.8\% RMSE reduction in Task 4). This suggests that hierarchically learning multi-scale spatial representations benefits more for coarse scales. Besides, \textit{One4All-ST (w/o SN)} performs much worse than \textit{One4All-ST}, especially for fine scales (RMSE doubling on Task 1 \& 2). This underscores the significance of employing a scale normalization layer, as applying the same transformation across all scales can disproportionately emphasize coarse scales, neglecting fine scales. Achieving a balance across all scales is crucial for modifiable areal units that may necessitate consideration at various scales.

\subsubsection{Effect of Hierarchical Structure}
As mentioned in Sec.~\ref{dataset_setup}, we construct the hierarchical structure using a $2\times 2$ merging window, resulting in 0.72M parameters. To explore how performance is affected by the merging window size (\textbf{RQ4}), we experiment with window sizes of $3\times 3$ and $4\times 4$, which produce hierarchical structures $\{1,3,9,27\}$ (0.54M parameters) and $\{1,4,16\}$ (0.46M parameters), respectively.

Fig.~\ref{fig: mws_analysis} shows results of different hierarchical structures based on \textit{One4All-ST}. In these structures, the $2\times 2$ variant performs the best, aligning with our expectations. The $2\times 2$ variant predicts more dense scales but requires more parameters than the other variants. However, on comparing the $3\times 3$ and $4\times 4$ variants, the $3\times 3$ variant generally underperforms, despite having more parameters and predicting additional scales. This discrepancy could stem from the greater importance of Scale 4 over Scale 3 for the given census tract and road segmentation queries. Another reason may be the extension of the atomic raster by zero-padding (to make the raster divisible to 3), incurring background noise to spatial representation learning for the $3\times 3$ variant.

The above analysis highlights the crucial role of hierarchical structures in modifiable areal unit prediction. In this paper, we opt for the 2$\times$2 merging window, achieving outstanding prediction performance with a modest computational cost (0.72M parameters). Furthermore, recognizing that region queries may prefer specific scales, we suggest a potential future direction for finding optimal hierarchical structures under resource-limited scenarios if region query scales could be pre-known.

\subsubsection{Analysis of Query Response Time} \label{sec: response_time}
To test our system's capability to handle online services for predicting arbitrary modifiable areal units, we display the response times of different tasks on Taxi NYC and Freight Transport datasets in Fig.~\ref{fig: response_time}. Recall that our system's online phase (Sec.~\ref{system_workflow}) involves two steps. Firstly, the deployed ST model continuously predicts crowd flow for all hierarchical grids and synchronizes with HBase at preset intervals (e.g., 1 hour). Then, region query prediction is achieved by retrieving and aggregating the optimal prediction of decomposed grids based on their index, eliminating the need for re-inferencing predictions. Therefore, we calculate the response time for region queries by summing decomposition time and indexing time. As shown in Fig.~\ref{fig: response_time}, the average response time increases with the task scale. Notably, the average response time for all four tasks on both datasets is below 2 milliseconds, with a maximum response time not exceeding 20 milliseconds. This already meets the requirements for online use.

\subsubsection{Effect of Spatial Modeling Block}
We compare the proposed \textit{One4All-ST} (SEBlock) with the variants using either the residual block (ResBlock) \cite{zhang2017deep} or the standard convolution block (ConvBlock) \cite{deepst_2016} as the spatial modeling block. As shown in Fig.~\ref{fig: effect_spatial_model_block}, SEBlock consistently outperforms ConvBlock and ResBlock by a significant margin in all cases, demonstrating the effectiveness of incorporating channel-wise information in feature maps and reducing MAPE up to 0.6\% compared to ResBlock, which is consistent with previous research findings \cite{STRN_2021}.

\subsubsection{Analysis of Index Size}
In Fig.~\ref{fig: index_size}, we calculate quad-tree index sizes in both datasets at each scale. The quad-tree indexes store optimal combinations for the hierarchical structure $P=\{1, 2, 4, 8, 16, 32\}$ with 150m$\times$150m atomic grids, supporting fine-grained ST prediction in a metropolis like Shanghai and NYC. The total index sizes for Taxi NYC and Freight Transport are relatively small at 66 MB and 64 MB respectively, making them suitable for loading into a single server for online region query processing.

\section{Related Work}
\subsection{Deep Learning for Spatio-Temporal Prediction}
Nowadays, deep neural networks are widely used for urban spatio-temporal prediction. Existing ST prediction models are categorized based on their input format into two main types: (i) \textit{grid-based models}, which divide the spatial domain evenly into $H\times W$ fine-grained mesh-grids and typically use a 4D tensor $\mathbb{R}^{T\times H\times W \times C}$ as input \cite{zhang2017deep, zhang_multi_2019}; and (ii) \textit{graph-based models}, which incorporate directed or undirected graphs to leverage the topological structure for modeling and usually take a 3D tensor $\mathbb{R}^{T\times N \times C}$ as input (where $N$ represents the number of nodes). Convolutional neural networks (CNN) are commonly used to capture local spatial dependencies in grid data, which is inherently Euclidean \cite{zhang2017deep, zhang_multi_2019, STRN_2021}. In contrast, except for Euclidean data, graphs are also well-suited for representing non-Euclidean data. Therefore, graph neural networks (GNNs) are beneficial for modeling irregular regions \cite{irregular_2022, flexible_partition_2020} and organizing a hierarchical structure with irregular partitions.  Additionally, GNNs can effectively capture long-range spatial dependencies using either predefined graphs \cite{li2017diffusion, chai_multi_graph_2018, Song_Lin_Guo_Wan_2020, GMAN_2020} or adaptive graphs \cite{graph_wavenet_2019, MTGNN_2020,dynamic_hypergraph_2023}.
In this paper, we use grids to create a hierarchical structure because grids do not have any prior preference for region partitioning (all grids are of equal size), making them suitable for arbitrary modifiable areal units.
Unlike most previous studies that make predictions on a specific scale, this paper proposes a framework that can give ST predictions for arbitrary modifiable areal units, which is a novel research problem and less studied in existing works. 

\subsection{Multi-scale Spatio-Temporal Learning}
In recent years, we have witnessed several pioneer work toward multi-scale spatio-temporal learning. Most of these focus on learning spatio-temporal representations on two scales (i.e., node level and cluster level). There are two main approaches to generating clusters: (i) \textit{feature-based methods}: cluster the nodes based on the features of nodes (e.g., historical traffic observations) \cite{SCEG_2020, MC_STGCN_2022}. For example, \textit{MC-STGCN} \cite{MC_STGCN_2022} performs both fine- and coarse-grained traffic flow predictions. The coarse-grained scale is clustered based on the topology information of the road network and historical traffic flow similarity. (ii) \textit{learning-based methods} construct clusters using learned node representations \cite{STRN_2021,HiSTGNN_2022}. For example, \textit{STRN} \cite{STRN_2021} predicts fine-grained urban flows by fusing coarse-grained cluster representations. The cluster is constructed based on high-level node representations extracted by the backbone network. Inspired by previous works, we build a hierarchical structure with different scale grids and design a more efficient network for multi-scale predictions (using fewer parameters and getting better results). More importantly, we study the optimal combination problem of how to leverage the multi-scale outputs for arbitrary modifiable areal unit prediction.


\section{Conclusion and Future Work}
This paper proposes \textit{One4All-ST}, which predicts spatio-temporal (ST) data for any modifiable areal unit using only one model, aiming at reducing the expensive cost and alleviating the prediction inconsistency caused by many ST models. 
To reduce the cost, we propose a hierarchical multi-scale ST network with spatial modeling and scale normalization modules to efficiently and equally learn multi-scale representations. To alleviate prediction inconsistencies, we propose a dynamic programming scheme to find the optimal combination with minimized predicted error for representing modifiable areal units. 
For in-time response in practical online scenarios, we introduce an extended quad-tree to index optimal combinations. Extensive experiments on real-world ST datasets demonstrate that \textit{One4All-ST} achieves the best prediction accuracy over competitive baselines with a much lower computation cost. 

Our future work would include: (1) we will develop approaches to determine the optimal hierarchical structure for further reducing computation costs in resource-limited scenarios. (2) we will utilize GNNs to improve long-range spatial dependencies modeling and explore hierarchical structures with irregular partitions that can be represented as graphs and modeled via GNNs.


\bibliographystyle{IEEEtran}
\bibliography{ref}

\begin{thebibliography}{10}
\providecommand{\url}[1]{#1}
\csname url@samestyle\endcsname
\providecommand{\newblock}{\relax}
\providecommand{\bibinfo}[2]{#2}
\providecommand{\BIBentrySTDinterwordspacing}{\spaceskip=0pt\relax}
\providecommand{\BIBentryALTinterwordstretchfactor}{4}
\providecommand{\BIBentryALTinterwordspacing}{\spaceskip=\fontdimen2\font plus
\BIBentryALTinterwordstretchfactor\fontdimen3\font minus \fontdimen4\font\relax}
\providecommand{\BIBforeignlanguage}[2]{{%
\expandafter\ifx\csname l@#1\endcsname\relax
\typeout{** WARNING: IEEEtran.bst: No hyphenation pattern has been}%
\typeout{** loaded for the language `#1'. Using the pattern for}%
\typeout{** the default language instead.}%
\else
\language=\csname l@#1\endcsname
\fi
#2}}
\providecommand{\BIBdecl}{\relax}
\BIBdecl

\bibitem{urban_monitoring_2011}
F.~Calabrese, M.~Colonna, P.~Lovisolo, D.~Parata, and C.~Ratti, ``Real-time urban monitoring using cell phones: A case study in rome,'' \emph{IEEE Transactions on Intelligent Transportation Systems}, vol.~12, no.~1, pp. 141--151, 2011.

\bibitem{digital_mobility_2023}
H.~Xu, A.~Berres, S.~B. Yoginath, H.~Sorensen, P.~J. Nugent, J.~Severino, S.~A. Tennille, A.~Moore, W.~Jones, and J.~Sanyal, ``Smart mobility in the cloud: Enabling real-time situational awareness and cyber-physical control through a digital twin for traffic,'' \emph{IEEE Transactions on Intelligent Transportation Systems}, vol.~24, no.~3, pp. 3145--3156, 2023.

\bibitem{joint_demand_2021}
H.~Yuan, G.~Li, Z.~Bao, and L.~Feng, ``An effective joint prediction model for travel demands and traffic flows,'' in \emph{2021 IEEE 37th International Conference on Data Engineering (ICDE)}, 2021, pp. 348--359.

\bibitem{bike_stg_2022}
G.~Li, X.~Wang, G.~S. Njoo, S.~Zhong, S.-H.~G. Chan, C.-C. Hung, and W.-C. Peng, ``A data-driven spatial-temporal graph neural network for docked bike prediction,'' in \emph{2022 IEEE 38th International Conference on Data Engineering (ICDE)}, 2022, pp. 713--726.

\bibitem{sthan_2022}
S.~Ling, Z.~Yu, S.~Cao, H.~Zhang, and S.~Hu, ``Sthan: Transportation demand forecasting with compound spatio-temporal relationships,'' \emph{ACM Trans. Knowl. Discov. Data}, oct 2022.

\bibitem{MVSTGN_2023}
Y.~Yao, B.~Gu, Z.~Su, and M.~Guizani, ``Mvstgn: A multi-view spatial-temporal graph network for cellular traffic prediction,'' \emph{IEEE Transactions on Mobile Computing}, vol.~22, no.~5, pp. 2837--2849, 2023.

\bibitem{st_aware_2022}
R.-G. Cirstea, B.~Yang, C.~Guo, T.~Kieu, and S.~Pan, ``Towards spatio- temporal aware traffic time series forecasting,'' in \emph{2022 IEEE 38th International Conference on Data Engineering (ICDE)}, 2022, pp. 2900--2913.

\bibitem{roi_demand_2023}
Y.~Cui, S.~Li, W.~Deng, Z.~Zhang, J.~Zhao, K.~Zheng, and X.~Zhou, ``Roi-demand traffic prediction: A pre-train, query and fine-tune framework,'' in \emph{2023 IEEE 39th International Conference on Data Engineering (ICDE)}, 2023, pp. 1340--1352.

\bibitem{ssl_traffic_2023}
S.~Guo, Y.~Lin, L.~Gong, C.~Wang, Z.~Zhou, Z.~Shen, Y.~Huang, and H.~Wan, ``Self-supervised spatial-temporal bottleneck attentive network for efficient long-term traffic forecasting,'' in \emph{2023 IEEE 39th International Conference on Data Engineering (ICDE)}, 2023, pp. 1585--1596.

\bibitem{graph_wavenet_2019}
Z.~Wu, S.~Pan, G.~Long, J.~Jiang, and C.~Zhang, ``Graph wavenet for deep spatial-temporal graph modeling,'' in \emph{Proceedings of the 28th International Joint Conference on Artificial Intelligence}, ser. IJCAI'19.\hskip 1em plus 0.5em minus 0.4em\relax AAAI Press, 2019, p. 1907–1913.

\bibitem{GMAN_2020}
C.~Zheng, X.~Fan, C.~Wang, and J.~Qi, ``Gman: A graph multi-attention network for traffic prediction,'' \emph{Proceedings of the AAAI Conference on Artificial Intelligence}, vol.~34, no.~01, pp. 1234--1241, 2020.

\bibitem{MTGNN_2020}
Z.~Wu, S.~Pan, G.~Long, J.~Jiang, X.~Chang, and C.~Zhang, ``Connecting the dots: Multivariate time series forecasting with graph neural networks,'' in \emph{Proceedings of the 26th ACM SIGKDD International Conference on Knowledge Discovery \& Data Mining}, ser. KDD '20, 2020, p. 753–763.

\bibitem{STRN_2021}
Y.~Liang, K.~Ouyang, J.~Sun, Y.~Wang, J.~Zhang, Y.~Zheng, D.~Rosenblum, and R.~Zimmermann, ``Fine-grained urban flow prediction,'' in \emph{Proceedings of the Web Conference 2021}, 2021, p. 1833–1845.

\bibitem{STMeta}
L.~Wang, D.~Chai, X.~Liu, L.~Chen, and K.~Chen, ``Exploring the generalizability of spatio-temporal traffic prediction: Meta-modeling and an analytic framework,'' \emph{IEEE Transactions on Knowledge and Data Engineering}, vol.~35, no.~4, pp. 3870--3884, 2023.

\bibitem{geng2019spatiotemporal}
X.~Geng, Y.~Li, L.~Wang, L.~Zhang, Q.~Yang, J.~Ye, and Y.~Liu, ``Spatiotemporal multi-graph convolution network for ride-hailing demand forecasting,'' \emph{Proceedings of the AAAI Conference on Artificial Intelligence}, vol.~33, no.~01, pp. 3656--3663, 2019.

\bibitem{ST_MetaNet_2022}
Z.~Pan, W.~Zhang, Y.~Liang, W.~Zhang, Y.~Yu, J.~Zhang, and Y.~Zheng, ``Spatio-temporal meta learning for urban traffic prediction,'' \emph{IEEE Transactions on Knowledge and Data Engineering}, vol.~34, no.~3, pp. 1462--1476, 2022.

\bibitem{zheng_deepstd_2020}
C.~Zheng, X.~Fan, C.~Wen, L.~Chen, C.~Wang, and J.~Li, ``Deepstd: Mining spatio-temporal disturbances of multiple context factors for citywide traffic flow prediction,'' \emph{IEEE Transactions on Intelligent Transportation Systems}, vol.~21, no.~9, pp. 3744--3755, 2020.

\bibitem{yuan_functions_2012}
J.~Yuan, Y.~Zheng, and X.~Xie, ``Discovering regions of different functions in a city using human mobility and pois,'' in \emph{Proceedings of the 18th ACM SIGKDD International Conference on Knowledge Discovery and Data Mining}, 2012, p. 186–194.

\bibitem{sun_community_2016}
\BIBentryALTinterwordspacing
L.~Sun, X.~Ling, K.~He, and Q.~Tan, ``Community structure in traffic zones based on travel demand,'' \emph{Physica A: Statistical Mechanics and its Applications}, vol. 457, pp. 356--363, 2016. [Online]. Available: \url{https://www.sciencedirect.com/science/article/pii/S0378437116300346}
\BIBentrySTDinterwordspacing

\bibitem{wong2004modifiable}
D.~W. Wong, ``The modifiable areal unit problem (maup),'' in \emph{WorldMinds: geographical perspectives on 100 problems}.\hskip 1em plus 0.5em minus 0.4em\relax Springer, 2004, pp. 571--575.

\bibitem{openshaw1981modifiable}
S.~Openshaw, ``The modifiable areal unit problem,'' \emph{Quantitative geography: A British view}, pp. 60--69, 1981.

\bibitem{de2021multicriteria}
S.~C. de~Andrade, C.~Restrepo-Estrada, L.~H. Nunes, C.~A.~M. Rodriguez, J.~C. Estrella, A.~C.~B. Delbem, and J.~Porto~de Albuquerque, ``A multicriteria optimization framework for the definition of the spatial granularity of urban social media analytics,'' \emph{International Journal of Geographical Information Science}, vol.~35, no.~1, pp. 43--62, 2021.

\bibitem{GridTuner_2022}
J.~Jin, P.~Cheng, L.~Chen, X.~Lin, and W.~Zhang, ``Gridtuner: Reinvestigate grid size selection for spatiotemporal prediction models,'' in \emph{2022 IEEE 38th International Conference on Data Engineering (ICDE)}, 2022, pp. 1193--1205.

\bibitem{RegionGen_2023}
L.~Chen, J.~Fang, Z.~Yu, Y.~Tong, S.~Cao, and L.~Wang, ``A data-driven region generation framework for spatiotemporal transportation service management,'' in \emph{Proceedings of the 29th ACM SIGKDD Conference on Knowledge Discovery and Data Mining}, ser. KDD '23, 2023, p. 3842–3854.

\bibitem{fine_inference_incomplete_2023}
J.~Li, S.~Wang, J.~Zhang, H.~Miao, J.~Zhang, and P.~S. Yu, ``Fine-grained urban flow inference with incomplete data,'' \emph{IEEE Transactions on Knowledge and Data Engineering}, vol.~35, no.~6, pp. 5851--5864, 2023.

\bibitem{zhang2017deep}
J.~Zhang, Y.~Zheng, and D.~Qi, ``Deep spatio-temporal residual networks for citywide crowd flows prediction,'' in \emph{Thirty-first AAAI conference on artificial intelligence}, 2017.

\bibitem{MC_STGCN_2022}
S.~Wang, M.~Zhang, H.~Miao, Z.~Peng, and P.~S. Yu, ``Multivariate correlation-aware spatio-temporal graph convolutional networks for multi-scale traffic prediction,'' \emph{ACM Trans. Intell. Syst. Technol.}, vol.~13, no.~3, jan 2022.

\bibitem{MTGCN_DASFAA_2020}
F.~Wang, J.~Xu, C.~Liu, R.~Zhou, and P.~Zhao, ``Mtgcn: A multitask deep learning model for traffic flow prediction,'' in \emph{Database Systems for Advanced Applications: 25th International Conference, DASFAA 2020, Jeju, South Korea, September 24–27, 2020, Proceedings, Part I}, 2020, p. 435–451.

\bibitem{STDM_2020}
S.~Wang, J.~Cao, and P.~S. Yu, ``Deep learning for spatio-temporal data mining: A survey,'' \emph{IEEE Transactions on Knowledge and Data Engineering}, vol.~34, no.~8, pp. 3681--3700, 2022.

\bibitem{Hive}
``Apache hive,'' \url{https://hive.apache.org/}, accessed: February 23, 2024.

\bibitem{HBase}
``Apache hbase,'' \url{https://hbase.apache.org/}, accessed: February 23, 2024.

\bibitem{HiSTGNN_2022}
Y.~Ma, P.~Gerard, Y.~Tian, Z.~Guo, and N.~V. Chawla, ``Hierarchical spatio-temporal graph neural networks for pandemic forecasting,'' in \emph{Proceedings of the 31st ACM International Conference on Information \& Knowledge Management}, ser. CIKM '22, New York, NY, USA, 2022, p. 1481–1490.

\bibitem{deepst_2016}
J.~Zhang, Y.~Zheng, D.~Qi, R.~Li, and X.~Yi, ``Dnn-based prediction model for spatio-temporal data,'' in \emph{Proceedings of the 24th ACM SIGSPATIAL International Conference on Advances in Geographic Information Systems}, ser. SIGSPACIAL '16, 2016.

\bibitem{swinTrans_2021}
Z.~Liu, Y.~Lin, Y.~Cao, H.~Hu, Y.~Wei, Z.~Zhang, S.~Lin, and B.~Guo, ``Swin transformer: Hierarchical vision transformer using shifted windows,'' in \emph{2021 IEEE/CVF International Conference on Computer Vision (ICCV)}, oct 2021, pp. 9992--10\,002.

\bibitem{bi2023accurate}
K.~Bi, L.~Xie, H.~Zhang, X.~Chen, X.~Gu, and Q.~Tian, ``Accurate medium-range global weather forecasting with 3d neural networks,'' \emph{Nature}, vol. 619, no. 7970, pp. 533--538, 2023.

\bibitem{SENet2018}
J.~Hu, L.~Shen, and G.~Sun, ``Squeeze-and-excitation networks,'' in \emph{2018 IEEE/CVF Conference on Computer Vision and Pattern Recognition}, 2018, pp. 7132--7141.

\bibitem{FPN_2017}
T.-Y. Lin, P.~Dollár, R.~Girshick, K.~He, B.~Hariharan, and S.~Belongie, ``Feature pyramid networks for object detection,'' in \emph{2017 IEEE Conference on Computer Vision and Pattern Recognition (CVPR)}, 2017, pp. 936--944.

\bibitem{spatial_trans_vis}
S.~Arlinghaus and J.~Kerski, ``Spatial transformations and visualization: Selected common threads and root concepts linking old to new,'' \emph{Solstice: An Electronic Journal of Geography and Mathematics}, vol. Volume XXV, 06 2015.

\bibitem{union_analysis}
``Union (analysis),'' \url{https://desktop.arcgis.com/en/arcmap/latest/tools/analysis-toolbox/union.htm}, accessed: February 23, 2024.

\bibitem{quad_r_tree_2002}
R.~K.~V. Kothuri, S.~Ravada, and D.~Abugov, ``Quadtree and r-tree indexes in oracle spatial: A comparison using gis data,'' in \emph{Proceedings of the 2002 ACM SIGMOD International Conference on Management of Data}, ser. SIGMOD '02, 2002, p. 546–557.

\bibitem{Il_quadtree_2013}
C.~Zhang, Y.~Zhang, W.~Zhang, and X.~Lin, ``Inverted linear quadtree: Efficient top k spatial keyword search,'' in \emph{2013 IEEE 29th International Conference on Data Engineering (ICDE)}, 2013, pp. 901--912.

\bibitem{overcome_ufi_2023}
H.~Yu, X.~Xu, T.~Zhong, and F.~Zhou, ``Overcoming forgetting in fine-grained urban flow inference via adaptive knowledge replay,'' in \emph{Proceedings of the Thirty-Seventh AAAI Conference on Artificial Intelligence and Thirty-Fifth Conference on Innovative Applications of Artificial Intelligence and Thirteenth Symposium on Educational Advances in Artificial Intelligence}, ser. AAAI'23/IAAI'23/EAAI'23, 2023.

\bibitem{uncertainty_quant_2023}
W.~Qian, D.~Zhang, Y.~Zhao, K.~Zheng, and J.~Q. Yu, ``Uncertainty quantification for traffic forecasting: A unified approach,'' in \emph{2023 IEEE 39th International Conference on Data Engineering (ICDE)}, apr 2023, pp. 992--1004.

\bibitem{dynamic_hypergraph_2023}
Y.~Zhao, X.~Luo, W.~Ju, C.~Chen, X.-S. Hua, and M.~Zhang, ``Dynamic hypergraph structure learning for traffic flow forecasting,'' in \emph{2023 IEEE 39th International Conference on Data Engineering (ICDE)}, 2023, pp. 2303--2316.

\bibitem{wang_functional_2023}
\BIBentryALTinterwordspacing
K.~Wang, L.~Liu, Y.~Liu, G.~Li, F.~Zhou, and L.~Lin, ``Urban regional function guided traffic flow prediction,'' \emph{Information Sciences}, vol. 634, pp. 308--320, 2023. [Online]. Available: \url{https://www.sciencedirect.com/science/article/pii/S0020025523004334}
\BIBentrySTDinterwordspacing

\bibitem{irregular_2022}
J.~Sun, J.~Zhang, Q.~Li, X.~Yi, Y.~Liang, and Y.~Zheng, ``Predicting citywide crowd flows in irregular regions using multi-view graph convolutional networks,'' \emph{IEEE Transactions on Knowledge and Data Engineering}, vol.~34, no.~5, pp. 2348--2359, 2022.

\bibitem{flexible_partition_2020}
X.~Wang, Z.~Zhou, Y.~Zhao, X.~Zhang, K.~Xing, F.~Xiao, Z.~Yang, and Y.~Liu, ``Improving urban crowd flow prediction on flexible region partition,'' \emph{IEEE Transactions on Mobile Computing}, vol.~19, no.~12, pp. 2804--2817, 2020.

\bibitem{CVNet2019}
X.~Tang, Z.~T. Qin, F.~Zhang, Z.~Wang, Z.~Xu, Y.~Ma, H.~Zhu, and J.~Ye, ``A deep value-network based approach for multi-driver order dispatching,'' in \emph{Proceedings of the 25th {ACM} {SIGKDD} International Conference on Knowledge Discovery {\&} Data Mining, {KDD} 2019, Anchorage, AK, USA, August 4-8, 2019}, 2019.

\bibitem{yuan2012segmentation}
N.~J. Yuan, Y.~Zheng, and X.~Xie, ``Segmentation of urban areas using road networks,'' Tech. Rep. MSR-TR-2012-65, July 2012.

\bibitem{xgboost_2016}
T.~Chen and C.~Guestrin, ``Xgboost: A scalable tree boosting system,'' in \emph{Proceedings of the 22nd ACM SIGKDD International Conference on Knowledge Discovery and Data Mining}, ser. KDD '16, 2016, p. 785–794.

\bibitem{zhang_multi_2019}
J.~Zhang, Y.~Zheng, J.~Sun, and D.~Qi, ``Flow prediction in spatio-temporal networks based on multitask deep learning,'' \emph{IEEE Transactions on Knowledge and Data Engineering}, vol.~32, no.~3, pp. 468--478, 2020.

\bibitem{li2017diffusion}
Y.~Li, R.~Yu, C.~Shahabi, and Y.~Liu, ``Diffusion convolutional recurrent neural network: Data-driven traffic forecasting,'' in \emph{International Conference on Learning Representations (ICLR '18)}, 2018.

\bibitem{chai_multi_graph_2018}
D.~Chai, L.~Wang, and Q.~Yang, ``Bike flow prediction with multi-graph convolutional networks,'' in \emph{Proceedings of the 26th ACM SIGSPATIAL International Conference on Advances in Geographic Information Systems}, ser. SIGSPATIAL '18, 2018, p. 397–400.

\bibitem{Song_Lin_Guo_Wan_2020}
C.~Song, Y.~Lin, S.~Guo, and H.~Wan, ``Spatial-temporal synchronous graph convolutional networks: A new framework for spatial-temporal network data forecasting,'' \emph{Proceedings of the AAAI Conference on Artificial Intelligence}, vol.~34, no.~01, pp. 914--921, Apr. 2020.

\bibitem{SCEG_2020}
Q.~Wang, B.~Guo, Y.~Ouyang, K.~Shu, Z.~Yu, and H.~Liu, ``Spatial community-informed evolving graphs for demand prediction,'' in \emph{Machine Learning and Knowledge Discovery in Databases. Applied Data Science and Demo Track: European Conference, ECML PKDD}, 2020, p. 440–456.

\end{thebibliography}

\end{document}